\documentclass{ecai}  

\usepackage{graphicx}
\usepackage{latexsym}

\paperid{1014}        

\usepackage{amsmath}
\usepackage{amsthm} 
\newtheorem{thm}{Theorem}

\newtheorem{definition}{Definition}

\usepackage{booktabs}

\usepackage{algorithm}
\usepackage{algorithmic}

\usepackage{comment}
\usepackage[nolist]{acronym}
\begin{acronym}
\acro{MSSS}{masked scaled standard simplex}
\acrodefplural{MSSS}{masked scaled standard simplices}

\acro{PSS}{padded standard simplex}
\acrodefplural{PSS}{padded standard simplices}

\acro{MDP}{Markov decision process}
\acrodefplural{MDP}{Markov decision processes}
\acro{CMDP}{constrained Markov decision process}
\acrodefplural{CMDP}{constrained Markov decision processes}
\acro{CRL}{constrained Reinforcement Learning}

\acro{MPT}{modern portfolio theory}
\acro{QP}{quadratic program}
\acro{DGP}{data generating process}
\acro{MV}{mean-variance}
\acro{RL}{reinforcement learning}
\acro{CVaR}{conditional value-at-risk}
\acro{AIC}{Akaike information criterion}
\acro{AIC}{Akaike information criterion}
\acro{BIC}{Bayesian information criterion}
\acro{HMM}{hidden Markov model}
\acro{MOO}{multi-objective optimization}
\acro{MOMDP}{multi-objective Markov decision process} 
\acro{ADBO}{Action Space Decomposition Based Optimization}

\acro{CAOSD}{Constrained Allocation Optimization with Simplex Decomposition}

\acro{ESM}{Environmental Score Metric}
\acro{MIQCP}{Mixed Integer Quadratically Constrained Program}

\end{acronym}
\usepackage{amssymb}
\usepackage[skip=0pt]{subcaption}

\usepackage{bbm}
\usepackage{dsfont}
\usepackage{xcolor}

\usepackage{enumitem}

\usepackage{array, booktabs, ragged2e}
\newcolumntype{L}[1]{>{\raggedright\let\newline\\\arraybackslash\hspace{0pt}}m{#1}}
\newcolumntype{C}[1]{>{\centering\let\newline\\\arraybackslash\hspace{0pt}}m{#1}}
\newcolumntype{R}[1]{>{\raggedleft\let\newline\\\arraybackslash\hspace{0pt}}m{#1}}

\begin{document}

\newcommand{\qedwhite}{\hfill \ensuremath{\Box}}

\newcommand{\setorig}{V}
\newcommand{\setorigsupport}{Q}
\newcommand{\setorigintersect}{\setorig_1 \cap \setorig_2}

\newcommand{\setconddecomp}{Q}

\newcommand{\constraintorig}{c}
\newcommand{\variableorig}{x}
\newcommand{\setall}{I}

\newcommand{\setsimplex}{K}
\newcommand{\constraintsimplex}{z}
\newcommand{\variablesimplex}{y}

\newcommand{\pssindexset}{K}
\newcommand{\pssonly}[1]{PSS_{\pssindexset_{#1}}}
\newcommand{\pssweighted}[1]{(\pssonly{#1})_{z_{#1}}}

\newcommand{\msssonly}[1]{MSSS_{\setsimplex_{#1},\constraintsimplex_{#1}}}
\newcommand{\msssonlydependent}[2]{MSSS_{\setsimplex_{#1},\constraintsimplex_{#1|#2}}}

\newcommand{\msssintro}[1]{}
\newcommand{\msssintrodependent}[1]{}
\newcommand{\TODO}[1]{{\color{red}TODO: {#1}}}

\newcommand{\PPOLagrange}{PPO-Lagrange} 
\newcommand{\LiuAL}{Liu et al. (2020)} 
\newcommand{\ZhangAL}{Zhang et al. (2022)} 
\newcommand{\TesslerAL}{Tessler et al. (2018)} 

\newcommand{\IPOmethod}{IPO method} 

\newcommand{\RCPO}{RCPO}
\newcommand{\IPO}{IPO}
\newcommand{\PTHREEO}{P3O}
\newcommand{\BASELINE}{Baseline}
\newcommand{\BASELINESHORT}{BL}
\newcommand{\RANDOM}{random}
\newcommand{\RANDOMSHORT}{RDM}
\newcommand{\CAOSD}{CAOSD}

\newcommand{\nasdaq}{Nasdaq-100}
\newcommand{\totaleconreturn}{\nu}
\newcommand{\estavgtotaleconreturn}{\bar{\totaleconreturn}}
\newcommand{\tableenv}{env}

\newcommand{\matprSG}{mean annualized total portfolio return}
\newcommand{\matprPL}{mean annualized total portfolio returns}

\newcommand{\transfunc}{{\mathcal{P}}}

\newcommand{\stateset}{{\mathcal{S}}}
\newcommand{\observationset}{{\mathcal{O}}}

\newcommand{\actset}{{\mathcal{A}_{2C}}}

\newcommand{\actionset}{{\mathcal{A}}}
\newcommand{\actionsetconstrained}{\actionset}
\newcommand{\actionsetsurrogate}{\tilde{\actionset}}

\newcommand{\actionsurrogate}{\tilde{a}}
\newcommand{\actionconstrained}{a}

\newcommand{\policyconstrained}{\pi}
\newcommand{\policysurrogate}{\tilde{\pi}}

\newcommand{\economicreturnportfolio}{\Theta_{PF}}
\newcommand{\economicreturnportfoliorealized}{\vartheta_{PF}}

\newcommand{\economicreturnvector}{\Theta}
\newcommand{\economicreturnvectorrealized}{\vartheta}

\newcommand{\riskestfunc}{\hat{f}_{risk, \economicreturnportfolio}}
\newcommand{\riskfunc}{f_{risk, \economicreturnportfolio}}

\newcommand{\assetallocationset}{\mathcal{V}}
\newcommand{\wealthset}{\mathcal{W}}
\newcommand{\risktoleranceset}{\mathcal{T}}
\newcommand{\marketset}{\mathcal{U}}
\newcommand{\transactioncost}{tc}
\newcommand{\borrowingfee}{bf}
\newcommand{\E}{{\rm I\!E}}

\newcommand{\setupAshort}{(A) simulation}
\newcommand{\setupBshort}{(B) backtesting}
\newcommand{\setupA}{\setupAshort~setup}
\newcommand{\setupB}{\setupBshort~setup}

\newcommand{\numberexperiments}{100}

\begin{frontmatter}

\title{Simplex Decomposition for Portfolio Allocation Constraints in Reinforcement Learning
}

\author[A,B]{\fnms{David}~\snm{Winkel}\orcid{0000-0001-8829-0863}\thanks{Corresponding Author. Email: winkel@dbs.ifi.lmu.de}}
\author[A,B]{\fnms{Niklas}~\snm{Strauß}\orcid{0000-0002-8083-7323}}
\author[A,B]{\fnms{Matthias}~\snm{Schubert}\orcid{0000-0002-6566-6343}} 
\author[A,B]{\fnms{Thomas}~\snm{Seidl}\orcid{0000-0002-4861-1412}} 

\address[A]{LMU Munich, Germany}
\address[B]{Munich Center for Machine Learning (MCML)}

\begin{abstract}
Portfolio optimization tasks describe sequential decision problems in which the investor's wealth is distributed across a set of assets.
Allocation constraints are used to enforce minimal or maximal investments into particular subsets of assets to control for objectives such as limiting the portfolio's exposure to a certain sector due to environmental concerns. Although methods for \ac{CRL} can optimize policies while considering allocation constraints, it can be observed that these general methods yield suboptimal results.
In this paper, we propose a novel approach to handle allocation constraints based on a decomposition of the constraint action space into a set of unconstrained allocation problems. In particular, we examine this approach for the case of two constraints. For example, an investor may wish to invest at least a certain percentage of the portfolio into green technologies while limiting the investment in the fossil energy sector. 
We show that the action space of the task is equivalent to the decomposed action space, and introduce a new \ac{RL} approach \acs{CAOSD}, which is built on top of the decomposition. The experimental evaluation on real-world \nasdaq~data demonstrates that our approach consistently outperforms state-of-the-art \ac{CRL} benchmarks for portfolio optimization.
\end{abstract}
\end{frontmatter}

\section{Introduction}
Portfolio optimization tasks belong to the family of resource allocation tasks in which an actor needs to allocate the available resources over a set of choices in each time step. Technically, resource allocation tasks can be considered multi-step decision problems with a standard-simplex action space describing all possible allocation choices, e.g., the set of all possible portfolio allocations in a financial setting.
Policy gradient based \ac{RL} can be used to optimize stochastic policies over the corresponding simplex action space and thus, they are often used to optimize portfolio allocation agents. For example, \cite{winkel22risk} proposes to use PPO \cite{schulman2017proximal} in combination with a Dirichlet action distribution for risk-aware portfolio optimization.

In many real-world financial settings, investors set maximum and minimum allocation weights to certain groups of assets for their portfolio. These constraints might originate from their client's investment guidelines, restrictions posed by the regulator, or the investors' economic opinion. For example, an investor might need to consider sustainability aspects in addition to generating economic returns. In this setting, the investor might be required to invest a minimal amount of 30\% of the portfolio into green technologies and a maximum amount of 15\% into companies belonging to the fossil energy sector. Allocation constraints reduce the set of allowed actions for the agent within the simplex action space, constraining the action space to a \emph{subset of the original simplex action space} that can be described geometrically as a convex polytope. Unfortunately, directly modeling a suitable action distribution on this polytope, which can be used to formulate a parametrizable policy function, is inherently difficult. A viable alternative approach is \acf{CRL}, which penalizes constraint violations in order to teach the agent to avoid disallowed actions, e.g., \cite{liu2020ipo,zhang2022penalized,tessler2018reward}.
However, most of these approaches cannot \emph{guarantee} that no constraint will be violated, may exhibit unstable
training behavior, or produce suboptimal results, especially if more than a single allocation constraint is needed.

In this paper, we propose an alternative approach that decomposes the original constraint action space into unconstrained sub-action spaces, each containing a subset of the assets. The actions from these sub-action spaces are then combined back into the original action space using a weighted Minkowski sum. We exploit that any constraint requiring a maximum investment into a subset of assets is equivalent to requiring a minimum investment in the inverse subset of assets. This allows us to consider only constraints requiring a minimum allocation to asset groups. For the case of two allocation constraints, we decompose the action space into four sub-action spaces. The first sub-action space invests into the assets that are restricted by both allocation constraints. The second and third sub-action spaces ensure the fulfillment of each constraint after the allocations in the first sub-action space. The final sub-action space freely distributes the remaining funds which were not needed to fulfill the constraints.
The allocation of assets in each sub-action space can be parameterized using an unconstrained Dirichlet distribution. We employ PPO \cite{schulman2017proximal} to optimize the policy function over the joint distribution of the sub-action spaces. Our new approach \acs{CAOSD} ensures a tractable computation of the joint probability and gradients of the sub-action spaces through an auto-regressive architecture. Additionally, we use a transformer-based encoder of the current pricing structure of the market which is based on the recent price development.

We demonstrate the effectiveness of our novel approach in portfolio optimization tasks based on real-world \nasdaq~index data. The results show that our approach is able to generate significantly higher returns than state-of-the-art constraint \ac{RL} methods in the overwhelming majority of cases.

The main contributions of this paper are the following:
\begin{itemize}
    \item We introduce a novel decomposition for constrained simplex action spaces with two allocation constraints into several unconstrained sub-action spaces.
    \item We propose a new method named \acs{CAOSD} utilizing this decomposition to effectively apply standard \ac{RL} algorithms like PPO on a surrogate action space. 
    \item We demonstrate that \acs{CAOSD} is able to significantly outperform state-of-the-art \ac{CRL} benchmark approaches on real-world market data.
\end{itemize}

Our paper is structured as follows: In Section~\ref{sec:relatedwork}, we give an overview of the related work and continue in Section~\ref{sec:problemsetting} with a formal problem definition. Afterward, we introduce our decomposition and show how to utilize the decomposition for \ac{RL} in Section~\ref{sec:propapproach}. We proceed with an extensive experimental evaluation of our approach in Section~\ref{sec:experiments} before concluding the paper in Section~\ref{sec:conclusion}.

\section{Related Work}
\label{sec:relatedwork}
Portfolio optimization tasks with allocation constraints can be formulated as \acp{CMDP}~\cite{altman1999constrained} and policies can be optimized using \ac{CRL} approaches~\cite{liu2020ipo,zhang2022penalized,tessler2018reward}. Note that our novel decomposition is able to parameterize the constraint action space directly, and thus, the task can be formulated as an (unconstrained) \ac{MDP}, for which optimal policies can be found through standard \ac{RL} algorithms.

\acp{CMDP} have seen successful applications in fields such as network traffic \cite{hou2017optimization}, and robotics  \cite{gu2017deep,achiam2017constrained}.
Finance applications such as \cite{coqueret2020dirichlet,winkel22risk} use a scalarized objective function to maximize the return while penalizing for risk to perform a multi-objective, i.e. risk/return, portfolio optimization. The allocation constraints in our setting allow also to control for risk as they can be used to limit the investor's exposure to risky sub-markets. \cite{abrate2021continuous} combine the risk/return optimization while also considering allocation constraints. They enforce these constraints by using a penalizing \ac{CRL} approach stating that ``there is no straightforward way to design an actor-network so that all proposed actions are compliant''. Our approach tackles this challenge and provides a way to do so on an actor-network level without the need for a penalty term.

There are different approaches to identify optimal policies in a \ac{CMDP} setting. Penalty-based approaches include an additional penalty term into the objective function representing the constraints. An example is Lagrangian-based approaches such as \cite{tessler2018reward,bhatnagar2012online} that transform a constrained optimization problem into an unconstrained one by applying Lagrangian relaxation. A subsequent step solves a saddle point problem by optimizing the objective function and dynamically adjusting the penalty factor $\lambda$.

Alternative penalty-based approaches, such as \cite{liu2020ipo}, employ interior-point methods. Common drawbacks of penalty-based approaches are the need for additional hyperparameter tuning, expensive training loops, and the lack of guarantees for satisfying the constraints.
An alternative approach is based on defining Trust Regions \cite{achiam2017constrained}, which may produce constraint violations due to approximation errors. Furthermore, there are approaches based on prior knowledge \cite{dalal2018safe} which pretrain a model to predict simple one-step dynamics of the environment. Other approaches are based on the use of Lyapunov functions to solve \acp{CMDP} by projecting either the policy parameter or the action onto the set of feasible solutions induced by state-dependent linearized Lyapunov constraints \cite{chow2018lyapunov,chow2019lyapunov}. However, this approach can be computationally expensive and, in some cases, numerically intractable, especially as the action space grows larger \cite{chow2019lyapunov}.

Our method relies on a novel decomposition of the action space into sub-spaces. Thus, approaches \emph{factorizing the action spaces} are another important research area that is related to this work. In \cite{tavakoli2018action}, the authors introduce action branching, which divides the action space into independent sub-action spaces. In contrast, our problem involves modeling sub-action spaces that depend on each other. Auto-regressive approaches, which can model dependencies between sub-action spaces \cite{metz2017discrete,wei2018hierarchical,pierrot2021factored}, are another common method for the factorization of the action spaces. Unlike these works, our approach focuses on a novel decomposition of a \emph{constrained simplex action space} to make the optimization problem easier to solve using standard \ac{RL} methods.

\section{Problem Setting}
\label{sec:problemsetting}
An \ac{MDP} is a 5-tuple $(\stateset,\actionset,\transfunc,\mathcal{R},\gamma)$ where $\stateset$ represents the state space, $\actionset$ the set of available actions, $\transfunc$ the transition function describing the distribution over future states $s'$ given a state-action pair $(s,a)$, $\mathcal{R}$ is a reward function $r:\stateset \times \actionset \times \stateset \rightarrow \mathbb{R}$ and $\gamma$ is the discount factor.

For portfolio allocation tasks, $\actionset$ is defined as a simplex over a set of $N$ assets $\setall=\{0,\ldots, N-1\}$, i.e., $\actionset = \left\{a \in \mathbb{R}_{0,+}^{N} : \sum_{i =0}^{N-1} {a_i =1}\right\} $. In other words, $a_i$ describes the positive relative amount of capital assigned to the $i^{th}$ asset. 

An allocation constraint is defined by a subset $V \subseteq \setall$ of assets and a threshold value $c \in [0,1]$ and implies that 
$\sum_{j\in V} {a_j \geq c}$ for all allocations $a \in \actionset$. Thus, at least the amount $c$ of the available capital must be allocated to assets from the set $V$.
Let us note that any \emph{less-than} constraint can be rewritten into a greater-equal constraint and vice versa. In particular, assigning \emph{at most} $c$ into assets from $V$ is equivalent to assigning \emph{more than} $(1-c)$ into the remaining assets $\setall \setminus V$, e.g., in a three asset setup with asset weights $x_1+x_2+x_3=1$ the greater-equal constraint $x_1+x_2 \geq 0.3$ is equivalent to the constraint $x_3 < 0.7$. 
Thus, without the loss of generality, we will assume greater-equal constraints in the following. For portfolio allocation tasks with one or more allocation constraints, the action space is a convex polytope within the original simplex of all $N$-dimensional allocation vectors.

The goal of a constrained portfolio allocation task is to find a policy $\pi_\theta$ maximizing the expected reward $\mathbb{E}_{\tau \sim \pi_\theta}[\sum_{t=0}^{T-1} {\gamma^t \cdot r_{t+1}}]$ where $r_{t+1}$ is the direct reward observed for the $t^{th}$ state transition of episodes $\tau$ sampled by $\pi_\theta$ while only using allowed allocations. In our setting, the reward corresponds to the direct economic returns of the entire portfolio.
Finding a suitable formulation for $\pi_\theta$ becomes more and more complex with an increasing number of allocation constraints. In the following, we will formulate $\pi_\theta$ for up to two allocation constraints which can directly be used in combination with standard actor-critic and policy gradient \ac{RL} algorithms. 
Thus, we consider the following action space with $V_1 \subseteq I$ and $V_2 \subseteq I$:
\begin{align*}
    \actset =& \biggl\{a \in \mathbb{R}_{0,+}^{N} : \sum_{i \in I}a_i =1,\sum_{j\in V_{1}} a_j \geq c_{1},\\
    & \sum_{k \in V_{2}} a_k \geq c_{2}, 0 \leq c_{1} \leq 1, 0 \leq c_{2} \leq 1 \biggr\}\\
\end{align*}

\section{\acf{CAOSD}}
\label{sec:propapproach}
As mentioned in Section \ref{sec:problemsetting}, the action space underlying our problem is a convex polytope within the standard simplex over a universe of $n$ assets. Thus, directly defining a parameterizable probability distribution that could be used in a policy function for reinforcement learning is rather complex. To avoid this complexity, we propose to decompose the action space into four standard simplices over subsets of $I$. For a proper weighting, allocations taken from these four standard simplices add up to form a complete allocation action in the original action space. In the following, we will describe the simplices and how to compute weights, guaranteeing that the original and the decomposed action set are equivalent. Afterward, we will describe how a policy function can be defined on top of the decomposed action space and how policy gradient based reinforcement learning methods can be applied.

\subsection{Action Space Simplex Decomposition}
\label{subsec:actionspacedecomp}
To formalize our approach, we begin by defining the basic elements of our decomposition, i.e., the padded standard simplices and their combination with the weighted Minkowski sum.

\begin{definition}
\label{def:pss}
Let $I=\{0,\ldots, N-1\}$ be a set of indices referring to respective dimensions in $\mathbb{R}^N$.
Let $S_{K}$ be a standard simplex in the subspace defined over the dimensions indicated by the index set $K \subseteq I$.
Let $g_K: \mathbb{R}^{|K|}\rightarrow \mathbb{R}^N$ be a function that projects $S_{K}$ into $\mathbb{R}^N$, by padding the entries for any elements in $\mathbb{R}^N$ in those dimensions with indices $I\setminus K$ with 0. Applying the function $g$ on $S_{K}$ then yields a \textbf{\acf{PSS}} defined as:
\begin{align*}
PSS_{K}=g_V(S_{K})=&\biggl\{ y \in \mathbb{R}^{N}_{0,+} : \sum_{j \in K} y_j = 1; y_i \geq 0 \: \forall i \in K;\\
&y_j =0 \quad \forall j \in I\setminus K \biggr\} \quad \text{ for } K\neq \emptyset
\end{align*}
and
\begin{align*}
PSS_{K}=g_V(S_{K}) = &\biggl\{ y \in \mathbb{R}^{N}_{0,+} : y_j=0 \quad \forall j \in I\biggr\}  \quad \text{ for } K= \emptyset
\end{align*}
\end{definition}

\begin{definition}
\label{def:weightedminkowskisum}
Given $n$ sets of vectors $Q_1,\ldots, Q_n$ in $\mathbb{R}^N$, the \textbf{weighted Minkowski sum} of $Q_1,\ldots, Q_n$ is generated by adding each combination of vectors from sets $Q_i$ after applying a respective weighting factor $z_i$, i.e., $(Q_i)_{z_i}=\{z_i\cdot q_i|q_i \in Q_i\}$  with $i=\{1,\ldots,n\}$. We write the weighted Minkowski sum of the sets as $M_z=(Q_1)_{z_1}+\ldots+(Q_n)_{z_n} = \{z_1\cdot q_1 + \ldots +z_n\cdot q_n | q_1\in Q_1, \ldots, q_n \in Q_n\}$.  We refer to $(Q_i)_{z_i}$ for $i=\{1,\ldots,n\}$ as the weighted Minkowski summands.
\end{definition}

Our approach identifies four \acp{PSS} and a weighting vector $z=[z_1,z_2,z_3,z_4]$ that can be combined as a weighted Minkowski sum $M_z = (PSS_1)_{z_1}+(PSS_2)_{z_2}+(PSS_3)_{z_3}+(PSS_4)_{z_4}$ such that $M_z=\actset$. 

\begin{figure}[t]
    \centering
    \includegraphics[width=0.33\textwidth]{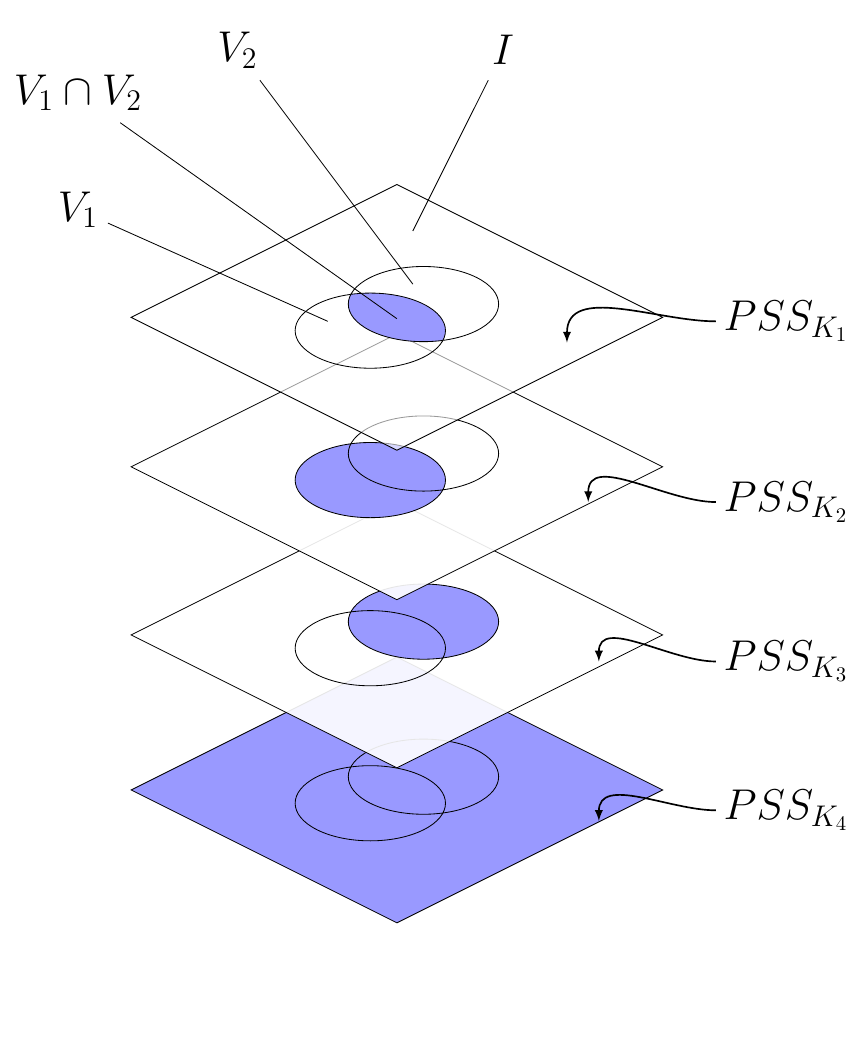}
    \caption{Set of padded variables represented as white area, set of modeled variables represented as colored area for each of the four \acp{PSS}.}
    \label{fig:setslayered}
\end{figure}

The first weighted Minkowski summand $\pssonly{1}$ is built over the intersection of assets $K_1= V_1 \cap V_2$. Investing into these assets contributes to fulfilling both constraints. In fact, if $c_1+c_2>1$, we need to invest at least a portion of $z_1=c_1+c_2-1$ into assets from $V_1 \cap V_2$ to avoid over investment. 

The second weighted Minkowski summand $\pssonly{2}$ is defined over the assets in $V_1=K_2$. To fulfill the first constraint, we have to make sure that we at least invest $c_1$ into assets in $V_1$. However, we need to consider that any capital $z_1$ already being allocated into $\pssonly{1}$ also contributes towards fulfilling this constraint. Correspondingly, $\pssonly{3}$ is defined over the assets in $K_3=V_2$ and requires an investment of $c_2$ minus any allocation made to $V_1 \cap V_2$ from $\pssonly{1}$ and $\pssonly{2}$. 
Finally, $\pssonly{4}$ is defined over the complete asset universe $I$. It covers the case, that not any available capital is needed to fulfill the allocation constraints. Thus, any remaining capital $1-(z_1+z_2+z_3)$ can be freely allocated among the assets in $I$ to maximize the economic return. An illustration of the four sets being covered by these weighted Minkowski summands can be found in Figure~\ref{fig:setslayered}.

To demonstrate the basic principle, consider an allocation task in which our capital must be allocated over five assets $a_1,a_2,a_3, a_4$ and $a_5$. The first constraint $c_1$ requires to allocate at least 30\% weight to the group of variables with index $V_1=\{1,3\}$. The second constraint $c_2$ requires to allocate at least 50\% weight to the group of variables with index $V_2=\{2,4\}$. Thus, the set of feasible solutions, i.e. the action space, is defined by the polytope 
\begin{align*}
    P_0 =& \biggl\{a\in \mathbb{R}^5: \sum_{i \in I}a_i=1; \sum_{i \in V_1}a_i \geq 0.3; \sum_{i \in V_2}x_i \geq 0.5;\\ 
    &x_i\geq 0 \quad \forall i \in I=\{1,2,3,4,5\}\biggl\}
\end{align*}

Given the four $\pssonly{j}$ with the respective index sets of $K_1 = \emptyset$, $K_2 = V_1$, $K_3=V_2$, $K_4=\setall$ and the weighting vector $z=[z_1, z_2, z_3, z_4] = [0.0, 0.3, 0.5, 0.2]$. The corresponding weighted Minkowski sum $M$ will equal $P_0$ as shown in the following:
Any final allocation $a = [a_1,a_2,a_3,a_4, a_5] \in M$ is the vector sum of four vectors $\actionsurrogate_j =[\actionsurrogate_{1,j},\actionsurrogate_{2,j},\actionsurrogate_{3,j},\actionsurrogate_{4,j},\actionsurrogate_{5,j}]\in \pssweighted{j} \subset \mathbb{R}^5$ for $j\in \{1,2,3,4\}$, i.e. $a=\actionsurrogate_1+\actionsurrogate_2+\actionsurrogate_3+\actionsurrogate_4$. 
The first sub-weighting vector $\actionsurrogate_1$ will be $(0,0,0,0,0)$ due to $K_1=\emptyset$ as we cannot invest into any assets. Any sub-weighting vector $\actionsurrogate_2 \in \pssweighted{2}$ will allocate a total of $z_2=0.3$ weight to the variables with an index in $V_1$, ensuring that any $a \in M$ will always satisfy the \emph{lower bound} of the first constraint $c_1=0.3$.
Any vector $\actionsurrogate_3 \in \pssweighted{3}$ will allocate a total of $z_3=0.5$ weight to the variables with an index in $V_2$, ensuring that any $a \in M$ will always satisfy the \emph{lower bound} of the second constraint $c_2=0.5$.
Any vector $\actionsurrogate_4 \in \pssweighted{4}$ will allocate the remainder of $z_4=0.2$ weight to any combination of variables in $I$, (a) ensuring that $\sum_{i\in I}a_i=1$ and (b) potentially allocating additional weight to the variables with indices in $V_1$ and $V_2$ (since $V_1 \subseteq I$ and $V_2 \subseteq I$), allowing $a \in M$ to exceed the lower bounds of $c_1$ and $c_2$, i.e. $\sum_{i \in V_1}a_i \geq 0.3$ and $\sum_{i \in V_2}a_i \geq 0.5$. 
As a result, the sets $M$ and $P_0$ have an identical H-representation (see Definition~\ref{def:convexpolytopehrepresentation}), i.e. being specified by the identical sets of constraints, making them identical polytopes.

In the following, we will introduce a weighting scheme selecting $z$ which guarantees general equivalence between $M$ and $P_0$. First, we formalize our constrained action space $A_{2C}$ as convex polytope and introduce its H-representation.

\begin{definition}
\label{def:convexpolytopehrepresentation}
A convex polytope $P$ in $\mathbb{R}^n$ is defined as a polytope that additionally is also a convex set. $P$ can be viewed as the set of solutions to a system of linear inequalities, i.e., the intersection of a finite number of closed half-spaces, called $P$'s half-space representation (H-representation):
\begin{align*}a_{{11}}x_{1}&&\;+\;&&a_{{12}}x_{2}&&\;+\cdots +\;&&a_{{1n}}x_{n}&&\;\geq \;&&&b_{1}\\a_{{21}}x_{1}&&\;+\;&&a_{{22}}x_{2}&&\;+\cdots +\;&&a_{{2n}}x_{n}&&\;\geq \;&&&b_{2}\\\vdots \;\;\;&&&&\vdots \;\;\;&&&&\vdots \;\;\;&&&&&\;\vdots \\a_{{m1}}x_{1}&&\;+\;&&a_{{m2}}x_{2}&&\;+\cdots +\;&&a_{{mn}}x_{n}&&\;\geq \;&&&b_{m}
\end{align*}
\end{definition}

The general formulation on how to identify the four $\pssonly{j}$ with their respective index sets $K_j$ and the definition of the weighting vector $z$ can be found in Theorem~\ref{theorem:generalcase}. 
Let us note that we define the weighting vector $z$ as an autoregressive function with $z=[z_1,z_2(z_1), z_3(z_1,z_2,y_2), z_4(z_1,z_2,y_2, z_3)]$ which results in an adaptive weighting vector depending on each current combination of elements in $\pssonly{j}$ for all $j=\{1,2,3,4\}$.

\begin{thm}
\label{theorem:generalcase}
    Any polytope $P$ defined by the system  
$$\sum_{i \in \setall} \variableorig_i = 1; \quad
\variableorig_i \geq 0 \quad \forall i\in\setall; \quad \sum_{i \in \setorig_1} \variableorig_i \geq \constraintorig_1; \quad \sum_{i \in \setorig_2} \variableorig_i \geq \constraintorig_2$$
with $\setall=\{0, \ldots, N-1\}$, $\setorig_1\subseteq I$ and $\setorig_2\subseteq I$ can be expressed as a weighted Minkowski sum with the four weighted Minkowski summands with $\variablesimplex_{i,j}=[\variablesimplex_{0,j},\ldots,\variablesimplex_{N-1,j}] \in \pssweighted{j}$:
    \begin{align*}
\pssweighted{1}:~& \setsimplex_1=\setorigintersect \text{ and } z_1=\max(0, \constraintorig_1+\constraintorig_2-1)\\
    \pssweighted{2}:~& \setsimplex_2=\setorig_1 \text{ and } \constraintsimplex_2 = \max(0,\constraintorig_1-z_1)\\
\pssweighted{3}:~& \setsimplex_3=\setorig_2 \text{ and } \constraintsimplex_3 = \max(0,\constraintorig_2-z_1-z_{2,\cap}) \\ & \text{ where } z_{2,\cap}=\sum_{i\in \setorig_1 \cap \setorig_2}y_{i,2}\\
\pssweighted{4}:~& \setsimplex_4=\setall \text{ and } \constraintsimplex_4 = 1-\constraintsimplex_1-\constraintsimplex_2-\constraintsimplex_3\\
\end{align*}
\end{thm}

\begin{proof}
\label{proof:theoremone}
Showing that two convex polytopes have an equivalent $H$-representation, i.e. an equivalent system of linear inequalities describing them, proves that they are identical.

When calculating the weighted Minkowski sum $M$ of the four \acp{PSS}, we can deduce that for any element $x = (\variablesimplex_1 + \variablesimplex_2 + \variablesimplex_3+ \variablesimplex_4) \in M$ with $\variablesimplex_1 \in \pssweighted{1}$, $\variablesimplex_2 \in \pssweighted{2}$, $\variablesimplex_3 \in \pssweighted{3}$ and $\variablesimplex_4 \in \pssweighted{4}$ the following constraints are fulfilled:

The contribution to the variables $\sum_{\setorig_1} \variableorig_{i}$ by $\pssweighted{1}$ and $\pssweighted{2}$ will always be $\max(0,c_1+c_2-1)+\max(0, c_1-z_1)=c_1$ while  $\pssweighted{3}$ and $\pssweighted{4}$ can optionally contribute positive weight, resulting in
\begin{align*}
	\sum_{i \in \setorig_1} \variableorig_{i}  = & \underbrace{\sum_{i \in \setorig_1 \cap \setsimplex_1} \variablesimplex_{i,1}}_{=\constraintsimplex_1=\max(0, c_1+c_2-1)} + \underbrace{\sum_{i \in \setorig_1\cap \setsimplex_2} \variablesimplex_{i,2}}_{= \constraintsimplex_2 = \max(0, c_1 -\constraintsimplex_1)} + \underbrace{\sum_{i \in \setorig_1\cap \setsimplex_3} \variablesimplex_{i,3}}_{\geq 0} \\
    &+ \underbrace{\sum_{i \in \setorig_1\cap \setsimplex_4} \variablesimplex_{i,4}}_{\geq 0} \geq c_1.
\end{align*}
The contribution to the variables $\sum_{\setorig_2} \variableorig_{i}$ by $\pssweighted{1}$, $\pssweighted{2}$, $\pssweighted{3}$ will always be $z_1+z_{2,\cap}+\max(0, c_2-z_1-z_{2,\cap})=c_2$ while $\pssweighted{4}$ can optionally contribute positive weight, resulting in
\begin{align*}
	\sum_{i \in \setorig_2} \variableorig_{i}  = & \underbrace{\sum_{i \in \setorig_2 \cap \setsimplex_1} \variablesimplex_{i,1}}_{=\constraintsimplex_1} + \underbrace{\sum_{i \in \setorig_2\cap \setsimplex_2} \variablesimplex_{i,2}}_{=z_{2,\cap}} + \underbrace{\sum_{i \in \setorig_2\cap \setsimplex_3} \variablesimplex_{i,3}}_{=\max(0,\constraintorig_2-z_1-z_{2,\cap})} \\
    & + \underbrace{\sum_{i \in \setorig_1\cap \setsimplex_4} \variablesimplex_{i,4}}_{\geq 0} \geq c_2.
\end{align*}

The total weight contribution from all four $\pssweighted{j}$ for $j=\{1,2,3,4\}$ to all variables $\sum_{\setall} \variableorig_{i}$ will be always $\constraintsimplex_1+\constraintsimplex_2+\constraintsimplex_3+(1-\constraintsimplex_1-\constraintsimplex_2-\constraintsimplex_3)=1$, i.e.
\begin{align*}
    \sum_{i \in \setall} \variableorig_{i} = & \underbrace{\sum_{i \in \setall \cap \setsimplex_1} \variablesimplex_{i,1}}_{=\constraintsimplex_1} + \underbrace{\sum_{i \in \setall\cap \setsimplex_2} \variablesimplex_{i,2}}_{=\constraintsimplex_2} + \underbrace{\sum_{i \in \setall\cap \setsimplex_3} \variablesimplex_{i,3}}_{=\constraintsimplex_3}\\
    &+ \underbrace{\sum_{i \in \setall\cap \setsimplex_4} \variablesimplex_{i,4}}_{=\constraintsimplex_4 = 1-\constraintsimplex_1-\constraintsimplex_2-\constraintsimplex_3}  = 1.
\end{align*}

Additionally, we check the constraints for the single variables $\variableorig_i$ for $i\in\setall$. Since for $y_{j}=[y_{0,j},\ldots, y_{N-1,j}]$ with $j=\{1,2,3,4\}$ all single variables $y_{i,j}$  with $i\in \setall$ are defined to be greater equal than zero, it follows that
\begin{align*}
\variableorig_i = \underbrace{\variablesimplex_{i,1}}_{\geq 0} + \underbrace{\variablesimplex_{i,2}}_{\geq 0}+ \underbrace{\variablesimplex_{i,3}}_{\geq 0} + \underbrace{\variablesimplex_{i,4}}_{\geq 0}\geq 0 \quad \forall i \in \setall
 \end{align*}
 
which shows the equivalence of the two sets of convex closed half-spaces defining $P$ and $M$, which proofs that $P=M$.
\end{proof}

\subsection{Task optimization via Reinforcement Learning}
\label{subsec:taskoptimizationRL}

\begin{figure}[t]
     \centering
    \includegraphics[width=0.48\textwidth]{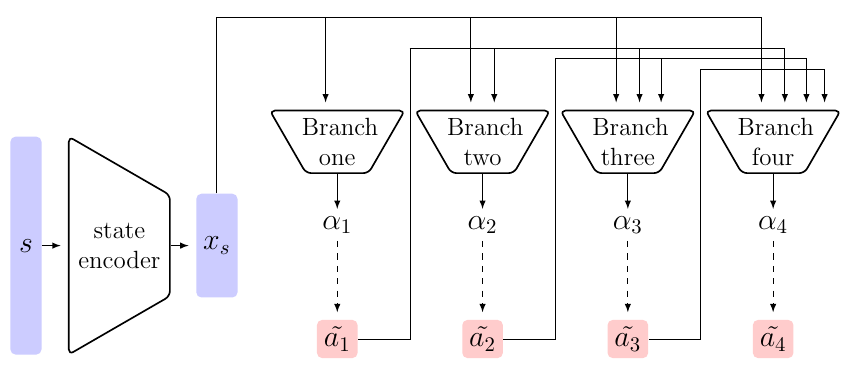}
\caption{Auto-regressive architecture. The dashed arrows represent the process of sampling from a Dirichlet distribution to generate a sub-action $\tilde{a}_j$.}
\label{fig:architectureautoregressive}
\end{figure}

After describing the simplex decomposition of the action space, we will now define a stochastic policy function based on our novel decomposition which can be used for policy optimization with policy gradient based \ac{RL} approaches.

We optimize the policy on a surrogate action space $\actionsetsurrogate=\actionsetsurrogate_1\times\actionsetsurrogate_2\times\actionsetsurrogate_3\times\actionsetsurrogate_4$, which is the Cartesian product of the four sub-action spaces. These sub-action spaces correspond to the four $\pssonly{j}$ when decomposing $\actset$ as introduced in the previous section. An action $\actionsurrogate \in \actionsetsurrogate$ can be mapped into the original action space by using the weighted asset-wise sum over the sub-action spaces: $a= z_1\cdot \actionsurrogate_{1}+z_2\cdot \actionsurrogate_{2}+z_3\cdot \actionsurrogate_{3}+z_4 \cdot \actionsurrogate_{4}$. 
Note that each surrogate action $\actionsurrogate$ maps to one particular action $a$ in the original action space $\actionset$ but not vice versa. In other words, the mapping is surjective but not bijective. Based on this property, we can show that any optimal policy on the surrogate action set $\actionsetsurrogate$ is optimal on the original action space $\actionset$ as well.
\begin{thm}
\label{theorem:optimality}
Given the constraint allocation task $(\stateset,\actionset,\transfunc,\mathcal{R},\gamma)$ as defined in Section~\ref{sec:problemsetting} with the surrogate action space $\actionsetsurrogate$ corresponding to the action decomposition defined in Theorem~\ref{theorem:generalcase}. Any optimal policy over the surrogate action space $\actionsetsurrogate$, $\pi^*_{\actionsetsurrogate}$ with $Q(s,\pi^*_{\actionsetsurrogate}(s)) \geq Q(s,\pi_{\actionsetsurrogate}(s))$ for any state $s \in \stateset$ and any policy $\pi_{\actionsetsurrogate}(s)$, implies an optimal policy $\pi^*_{\actionset}$ on the original action space $\actionset$.
\end{thm}
\begin{proof}
We know from Theorem \ref{theorem:generalcase} that there exist weights $z$ which allow to represent any $a \in \actionset$ by at least one surrogate action $\actionsurrogate$. In addition, we know that for any state $s \in \stateset$ and any action $a \in \actionset$, the state-action value function $Q(s,a)$ is the same for any $\actionsurrogate$ mapping to $a$ as the reward received for performing $\actionsurrogate$ is provided by the environment via performing the joint action $a$.

Thus, for any optimal policy $\pi^*_{\actionsetsurrogate}$, there exists a corresponding policy $\pi^*_{\actionset}$ providing the same Q-values. Now assume that $\pi^*_{\actionset}$ is not optimal and thus, there would be a policy $\hat{\pi}_{\actionset}$ with $Q(s,\hat{\pi}_{\actionset}(s))>Q(s,\pi^*_{\actionset}(s))$ for at least one state $s \in \stateset$. Since the mapping between the $\actionsetsurrogate$ and $\actionset$ is surjective, there must exist a decomposition of $\hat{\pi}_{\actionset}(s)$ to the surrogate action space yielding higher Q-values than $\pi^*_{\actionsetsurrogate}(s)$ which contradicts the optimality of $\pi^*_{\actionsetsurrogate}$.
\end{proof}
Theorem~\ref{theorem:optimality} shows that any well-performing policy in $\actionsetsurrogate$ can be mapped to an equally well-performing policy over $\actionset$. 

After introducing the surrogate action space, we will introduce our stochastic policy function over $\actionsetsurrogate$.
We model each sub-action space with a Dirichlet distribution with a parameter vector $\alpha_{K_j}$, which is obtained by a neural network. Our policy function that is based on the decomposition described in Theorem~\ref{theorem:generalcase}, requires an iterative computation of the weighting vector $z$. Our algorithm for sampling an action is detailed in Algorithm~\ref{alg:algorithm}. In each step, we sample the asset allocation for $\pssonly{j}$ by the corresponding Dirichlet distribution $Dir_j$ and then determine the corresponding weight $z_j$. In the end, the weighted asset-wise sum is computed and returned as joint action which is applied to the environment.

An overview of our architecture is depicted in Figure~\ref{fig:architectureautoregressive}. We first create a representation $x_s$ of the observation $s$ using a transformer model. Each sub-action space is parameterized by an MLP using the representation $x_s$ as well as all sampled surrogate actions from the previous sub-action spaces. 
This structure allows a tractable computation and optimization of the joint surrogate action $\actionsurrogate$ probability: $P(\actionsurrogate_1, \actionsurrogate_2, \actionsurrogate_3, \actionsurrogate_4 | x_s) = P(\actionsurrogate_1 | x_s) \cdot P(\actionsurrogate_2 | \actionsurrogate_1, x_s) \cdot P(\actionsurrogate_3 | \actionsurrogate_2, \actionsurrogate_1, x_s) \cdot P(\actionsurrogate_4 | \actionsurrogate_3, \actionsurrogate_2, \actionsurrogate_1, x_s)$.

We employ a policy gradient approach based on the PPO algorithm introduced by \cite{schulman2017proximal}. Note that our method can also be used with other policy gradient based RL methods.

The state encoder is composed of three fully connected layers of size $512$, $256$, and $128$ with ReLU activation functions that feed into a GTrXL element, allowing also to handle tasks that require memory. GTrXL is based on \cite{parisotto2020stabilizing} and is specifically designed to utilize transformers in an \ac{RL} setting. The GTrXL element is composed of a single \emph{transformer unit} with a single encoder layer and a single decoder layer with four attention heads and an embedding size of 64. The different branching modules are all made up of two fully connected layers of size $64$ and $32$, respectively, with a ReLU activation function after the first layer and a softplus activation function for the final output layer.

\begin{algorithm}[tb]
    \caption{Action Generation using the Simplex Decomposition}
    \label{alg:algorithm}
    \textbf{Input}: Index set of all $N$ assets in the investable universe $I=\{0,1,..., N-1\}$; Two allocation constraints $C_1: \sum_{i\in V_1}x_i\geq c_1$ and $C_2: \sum_{i\in V_2}x_i\geq c_2$\\
    \textbf{Define}:\\
    $K_1=\setorig_1 \cap \setorig_2$, $K_2=V_1$, $K_3=V_2$ and $K_4=I$, $f_j$ is an autoregressive policy network branch for $j=\{1,2,3,4\}$\\
    \textbf{Begin action generation:}
\begin{algorithmic}[1]
    \STATE calculate $\alpha_{1}$ from $f_1(x_s)$, sample $\Tilde{a}_1$ from $Dir(\alpha_{1})$
    \STATE set $z_{1}=\max(0, c_1+c_2-1)$
    \STATE set $q_{\setorigintersect}:=z_1\sum_{i\in \setorigintersect}x_{i, 1}$
    \STATE calculate $\alpha_{2}$ from $f_2(x_s, \Tilde{a}_1)$, sample $\Tilde{a}_2$ from $Dir(\alpha_{2})$
    \STATE set $z_{2}=\max[0,c_1-z_{1}]$
    \STATE update $q_{\setorigintersect}:=q_{\setorigintersect}+z_2\sum_{i\in \setorigintersect}x_{i, 2}$
    \STATE calculate $\alpha_{3}$ from $f_3(x_s, \Tilde{a}_1, \Tilde{a}_2)$, sample $\Tilde{a}_3$ from $Dir(\alpha_{3})$
    \STATE set $z_{3}=\max[0,c_2-q_{\setorigintersect}]$
    \STATE calculate $\alpha_{4}$ from $f_4(x_s, \Tilde{a}_1, \Tilde{a}_2, \Tilde{a}_3)$, sample $\Tilde{a}_4$ from $Dir(\alpha_{4})$
    \STATE set $z_{4}=1-z_{1}-z_{2}-z_{3}$
    \STATE calculate action $a$ by adding the weighted sub-actions:\\ $z_1\cdot \Tilde{a}_{1}+z_2\cdot \Tilde{a}_{2}+z_3\cdot \Tilde{a}_{3}+z_4 \cdot \Tilde{a}_{4} = a$
\end{algorithmic}
\end{algorithm}

\section{Experiments}
\label{sec:experiments}
\subsection{Constrained Portfolio Optimization Tasks}
We evaluate our approach in the financial setting on various constrained portfolio optimization tasks. The environment is based on \cite{winkel22risk} and uses real-world data from the \nasdaq~index  that has been processed by the qlib package.\footnote{https://github.com/microsoft/qlib/tree/main}
The data is used to estimate the parameters of a \ac{HMM}, which is then used to generate trajectories. The monthly closing stock prices from January 1, 2010 to December 31, 2020 are included in the data set. An additional data set containing monthly closing prices from January 1, 2021 to December 31, 2021 is exclusively used to backtest the approaches. 
The environment's investment universe consists of 12 assets plus the special asset cash. Cash has neither a positive or negative return and remains stable over time. The remaining 12 assets are chosen at random from a pool of 35 pre-selected assets from the \nasdaq~data set. The assets were pre-selected based on the fact that they have been a member of the index at least since January 1, 2010, and there were no missing data entries.

In the following, we provide a detailed description of the portfolio optimization task.
An agent is required to invest
his wealth into $N$ different assets based  on asset allocation decisions made at each time step  of the investment horizon $T$.
The \emph{constrained} \textbf{action space} for this task is described by $\actset$ as defined in Section~\ref{sec:problemsetting}, where the sets $V_1$ and $V_2$ contain the indices of assets affected by the respective constraint. 
These allocation constraints in the financial setting can be motivated by various factors, such as the need to invest \emph{at least} a certain percentage of the portfolio into a group of assets in a specific sector or with a certain risk classification.

The \textbf{observation space} is defined as $\observationset=\wealthset\times\assetallocationset\times\marketset$. The set $\wealthset \subseteq \mathbb{R}$ represents the agent's current absolute level of wealth, the set $\assetallocationset\subseteq\mathbb{R}^N$ describes the current portfolio allocation, and the set $\marketset\subseteq\mathbb{R}^N$ is the observed single asset economic returns from the previous time step.

The total portfolio return  $\label{eq:economicperformance} r = \economicreturnportfoliorealized - \transactioncost$ is the agent's \textbf{reward} for each time step and is defined as the realized portfolio's return $\economicreturnportfoliorealized$ minus any transaction costs $\transactioncost$ that occurred.
The portfolio's return is a random variable $\economicreturnportfolio = a^\intercal \economicreturnvector$ based on the random vector $\economicreturnvector~=[\economicreturnvector_{0},\ldots , \economicreturnvector_{N-1}]$ representing the economic returns of the single assets and the deterministic vector $a$ that represents the portfolio's allocation weights selected by the agent. The cumulative portfolio total return over the investment horizon of 12 months, i.e. $T=12$ time steps, is defined as 
\begin{align*}
    \totaleconreturn = \sum_{t=0}^{T-1}r_{t+1}
\end{align*}
and in the following referred to as the \emph{annualized total portfolio return}.

\subsection{Experimental Setup}

As previously stated, two evaluation environments are used: (1) the \emph{simulation environment} and (2) the \emph{backtesting environment}.
We conduct a total of \numberexperiments~experiments, each with a unique constraint configuration, i.e., a unique combination of two allocation constraints. Each constraint configuration is evaluated on both evaluation environments with the goal of comparing the performance of the approaches for different constraint configurations (a) on the environment the agents were originally trained on and (b) on \emph{unseen, real world} data.

Each experiment uses a different random seed as well as a randomly generated constraint configuration.
A constraint configuration 
is made up of two allocation constraints $C_j$ of the form $\sum_{i \in V_j} a_{i} \geq c_j$ with $j \in\{1,2\}$ where $V_j$ represents the set of affected assets and $c_j$ the constraint's threshold value.
To generate both allocation constraints at random, we sample the number of affected assets between 1 and 12. We rule out the possibility of selecting 0 or 13 assets since any greater equal allocation constraint would be either infeasible or trivial. The sampled number defines the number of specific assets which are then sampled from the list of 13 assets, i.e. the investment universe, without replacement resulting in $V_j$. Subsequently $c_j$ is sampled from a uniform distribution in the interval $[0,1]$.
The process is repeated for the second allocation constraint as well resulting in a randomly generated constraint configuration. In a final step it is verified that the resulting polytope $P$ as defined in Section~\ref{sec:problemsetting} is not an empty set, i.e. a system that does not have a feasible solution.\footnote{This can be checked by determining whether the V-representation of $P$ contains at least one vertex.}

We compare our \ac{CAOSD} approach to four other approaches, one of which is a naive \RANDOM~approach and three of which are state-of-the-art \ac{CRL} approaches. The \ac{CRL} approaches typically model constraint violations on a trajectory level, which means that they constrain the expected discounted sum of costs that occurred in each time step \cite{altman1999constrained}. However, they can be adjusted to model allocation constraints that must be satisfied at each time step. This can be done by defining the costs at each time step in such a way that they return a positive value if an allocation constraint is violated and zero otherwise. When a violation occurs in any time step, the discounted sum of costs will be greater than zero. Therefore, we can constrain every time step in the trajectory implicitly by imposing a constraint on the trajectory level that the expected discounted sum of costs needs to be less than or equal to zero.

The first \ac{CRL} benchmark approach is the Lagrangian-based \RCPO~introduced by \cite{tessler2018reward}. The second benchmark approach is \IPO~proposed by \cite{liu2020ipo} that uses an interior-point method to optimize the policy. The third approach is \PTHREEO~by \cite{zhang2022penalized}, a first-order optimization approach that uses an unconstrained objective in combination with a penalty term equaling the original constraint objective. The benchmark approaches are implemented in the RLlib framework based on their papers.\footnote{https://docs.ray.io/en/master/rllib/index.html} The code for all approaches is made publicly available.\footnote{https://github.com/DavWinkel/SimplexDecompositionECAI} All agents were trained on a cluster using various types of commercially available single GPUs. All approaches were extensively tuned in terms of hyperparameters using a grid search. Additional information on the hyperparameter tuning process can be found in the Appendix. During evaluation, \ac{RL} agents take the action with the highest likelihood.

In addition to the three benchmark approaches we also utilize a \RANDOM~approach. This approach uniformly draws actions, i.e. asset allocations, from the constrained polytope. Efficient uniform sampling from a polytope is a surprisingly complex task, therefore we follow \cite{corte2021novel} to obtain uniform samples from the constrained action space.
Using several rollouts of this baselines allows us to establish an estimate of the difficulty of an experiment, since the possible returns are highly dependent on the allocation constraints.

\subsection{Evaluation}
In our evaluation, we first compare the performance of our approach and the benchmarks over the entire set of experiments, which demonstrates the effectiveness of our approach in various settings. Afterward, we discuss the performance and convergence during training and take a detailed look at a single experiment. 
A key metric of the evaluation is the agents' \matprPL. We define the \matprSG~for each of the five approaches, i.e. $app=\{\RCPO, \IPO, \PTHREEO,\CAOSD, \RANDOMSHORT\}$, and each of the environments, i.e. $env=\{sim, bt\}$, as $\estavgtotaleconreturn^{env}_{app}=\dfrac{1}{J}\sum_{j=0}^{J-1}\totaleconreturn^{env}_{app,j}$ where $J$ is the number of evaluation trajectories. For the simulation environment $J=1000$ trajectories per approach are evaluated after the agents' training is completed.

In backtesting -- with the exception of the \RANDOM~approach -- only the single real-world trajectory is evaluated to measure the agents' performance since their evaluation is deterministic. The \RANDOM~approach is treated differently since its evaluation remains  stochastic due to its previously mentioned design to always sample uniformly an action from $P$. To reduce the variance in the results, we evaluate $\estavgtotaleconreturn^{bt}_{\RANDOMSHORT}$ on $J=1000$ rollouts during backtesting.

\begin{table}
	\centering
	\begin{tabular}{c C{1.6cm} C{1.6cm}  C{1.6cm}}
		\toprule
		& $\bar{\theta}^{env}_{app}$ & Upper 95\% CI & Lower 95\% CI\\[0.5ex]
		\midrule
		\textbf{Simulation} & & & \\
		RCPO& 0.292 & 0.299 & 0.284\\
        IPO& 0.212 & 0.217 & 0.207\\
        P3O& 0.305 & 0.314 & 0.296\\
        CAOSD& \textbf{0.327} & 0.335 & 0.319\\
        Random& 0.209 & 0.217 & 0.201\\
        \midrule
        \textbf{Backtesting} & & & \\
        RCPO& 0.497 & 0.521 & 0.474\\
        IPO& 0.35 & 0.365 & 0.334\\
        P3O& 0.522 & 0.552 & 0.492\\
        CAOSD& \textbf{0.552} & 0.582 & 0.522\\
        Random& 0.333 & 0.355 & 0.311\\
		\bottomrule
	\end{tabular}
	\caption{Evaluation of $\bar{\theta}^{env}_{app}$ and its 95\% confidence interval for all approaches in both environments for N=100 experiments after training is completed.}
	\label{tab:evaluationaverageinferenceabsoluteperf}
\end{table}
We use two measures to evaluate the performance of the approaches over all experiments. The first measure, $\bar{\theta}^{env}_{app}$, is the average of the mean annualized return of each approach over all experiments. More formally, $\bar{\theta}^{env}_{app} = \dfrac{1}{N} \sum_{i=0}^{N-1}\estavgtotaleconreturn^{env}_{app,i}$, where $N=\numberexperiments$ is the number of experiments. Since the return that can be achieved in each experiments varies greatly depending on the constraint configuration, our second measure is defined as the difference of returns of each approach to the performance of the \RANDOM~baseline in the same experiment: $\bar{\delta}^{env}_{app}=\dfrac{1}{N}\sum_{i=0}^{N-1}\estavgtotaleconreturn^{env}_{app,i}-\estavgtotaleconreturn^{env}_{\RANDOMSHORT,i}$.

\begin{table}
	\centering
	\begin{tabular}{c C{1.6cm} C{1.6cm}  C{1.6cm}}
		\toprule
		& $\bar{\delta}^{env}_{app}$ & Upper 95\% CI & Lower 95\% CI\\[0.5ex]
		\midrule
		\textbf{Simulation} & & & \\
		RCPO& 0.082 & 0.093 & 0.071\\
        IPO& 0.003 & 0.012 & -0.007\\
        P3O& 0.096 & 0.108 & 0.084\\
        CAOSD& \textbf{0.118} & 0.129 & 0.106\\
        \midrule
        \textbf{Backtesting} & & & \\
        RCPO& 0.164 & 0.196 & 0.132\\
        IPO& 0.017 & 0.046 & -0.012\\
        P3O& 0.189 & 0.227 & 0.151\\
        CAOSD& \textbf{0.219} & 0.258 & 0.179\\
		\bottomrule
	\end{tabular}
	\caption{Evaluation of $\bar{\delta}^{env}_{app}$ and its 95\% confidence interval for the non-random approaches in both environments for N=100 experiments after training is completed.}
	\label{tab:evaluationaverageinferencedeltabaseline}
\end{table}

Table~\ref{tab:evaluationaverageinferenceabsoluteperf} and Table~\ref{tab:evaluationaverageinferencedeltabaseline} show the performance of the approaches for both metrics in both environments as well as their corresponding 95\% confidence intervals. 
The \ac{CAOSD} approach shows considerable improvements over the other approaches in both metrics and both environments. These improvements are statistically significant on a 95\% confidence interval. The \PTHREEO~approach ranks second in both environments for both metrics before \RCPO. \IPO~is only able to outperform the \RANDOM~approach in the backtesting environment while producing similar performance results to the \RANDOM~approach in the simulation environment.

In the second part of the evaluation, we will discuss the performance of the agents \emph{during training} on a representative experiment. The experiment has a constraint configuration with the two allocation constraints $C_1: \sum_{i \in V_1}a_{i} \geq 0.23$ with $V_1$ containing the indices referring to the company stocks [BIDU, QCOM] and $C_2:\sum_{i \in V_2}a_{i} \geq 0.32$ with $V_2$ referring to the indices of the companies [ADBE, SBUX, QCOM] (see Appendix for a detailed list of the environment's investment universe).
During training, an evaluation with $J=200$ trajectories is performed every 80000 environment steps.

\begin{figure}[t]
    \centering
    \includegraphics[width=\columnwidth]{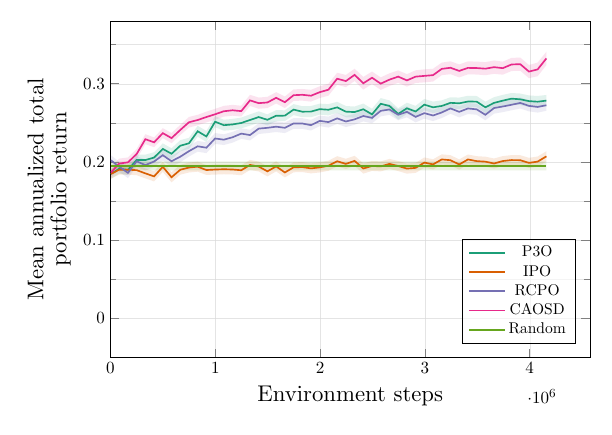}
    \caption{Mean annualized total portfolio return during training and its 95\%~confidence interval. This figure is best viewed in color.
    }
    \label{fig:meaneconreturnsingleexperiment}
\end{figure}

Figure~\ref{fig:meaneconreturnsingleexperiment} shows the agents' \matprSG~during training on the y-axis and the number of environment steps on the x-axis. The figure also shows the 95\% confidence interval of the \matprSG~for each of the approaches seen as the shaded areas around the lines. Note that we only show the \emph{training performance} for the simulation environment, since there is no training in the backtesting environment.
The \ac{CAOSD} approach had the best training performance in the experiment shown in Figure~\ref{fig:meaneconreturnsingleexperiment}, followed by \PTHREEO~and \RCPO. The \IPO~approach is not able to improve the \matprSG~during training and stays comparable to the \RANDOM~approach.

\section{Conclusion}
\label{sec:conclusion}
In this paper, we examine portfolio optimization tasks with allocation constraints that require investing at least a certain portion of the available capital into a subset of assets. The task covers many real-world use-cases such as investors wanting to limit their exposure to certain groups of assets due to risk concerns, or investors who want to reflect aspects such as sustainability or social responsibility in their portfolio allocation.
We examine settings that consider two allocation constraints and present \ac{CAOSD} which decomposes the constrained action space into multiple unconstrained sub-action spaces. We show that the weighted Minkowski sum of these sub-action spaces is equivalent to the original action space if weights are chosen properly. Based on the decomposition we introduce a stochastic policy function that computes proper weights with an auto-regressive pattern. To optimize the policy for a given task, we apply a transformer-based state encoder and employ PPO \cite{schulman2017proximal} to train our agent. In the experimental part, we apply our approach to a variety of constrained portfolio optimization tasks, each characterized by a different set of constraints. We significantly outperform state-of-the-art approaches from \ac{CRL} on real-world market data which demonstrates the effectiveness of our proposed method.

While this work shows decomposition with up to two allocation constraints, we will investigate decompositions for a greater number of constraints in future work. This increases the complexity of the possible relationship structures between the sets of choices, necessitating increasingly complex decompositions.
In another line of future work, we want to examine the application of our approach to other tasks than portfolio optimization.

\bibliography{ecai}
\end{document}


\newcommand{\qedwhite}{\hfill \ensuremath{\Box}}

\newcommand{\setorig}{V}
\newcommand{\setorigsupport}{Q}
\newcommand{\setorigintersect}{\setorig_1 \cap \setorig_2}

\newcommand{\setconddecomp}{Q}

\newcommand{\constraintorig}{c}
\newcommand{\variableorig}{x}
\newcommand{\setall}{I}

\newcommand{\setsimplex}{K}
\newcommand{\constraintsimplex}{z}
\newcommand{\variablesimplex}{y}

\newcommand{\pssindexset}{K}
\newcommand{\pssonly}[1]{PSS_{\pssindexset_{#1}}}
\newcommand{\pssweighted}[1]{(\pssonly{#1})_{z_{#1}}}

\newcommand{\msssonly}[1]{MSSS_{\setsimplex_{#1},\constraintsimplex_{#1}}}
\newcommand{\msssonlydependent}[2]{MSSS_{\setsimplex_{#1},\constraintsimplex_{#1|#2}}}

\newcommand{\msssintro}[1]{}
\newcommand{\msssintrodependent}[1]{}
\newcommand{\TODO}[1]{{\color{red}TODO: {#1}}}

\newcommand{\PPOLagrange}{PPO-Lagrange} 
\newcommand{\LiuAL}{Liu et al. (2020)} 
\newcommand{\ZhangAL}{Zhang et al. (2022)} 
\newcommand{\TesslerAL}{Tessler et al. (2018)} 

\newcommand{\IPOmethod}{IPO method} 

\newcommand{\RCPO}{RCPO}
\newcommand{\IPO}{IPO}
\newcommand{\PTHREEO}{P3O}
\newcommand{\BASELINE}{Baseline}
\newcommand{\BASELINESHORT}{BL}
\newcommand{\RANDOM}{random}
\newcommand{\RANDOMSHORT}{RDM}
\newcommand{\CAOSD}{CAOSD}

\newcommand{\nasdaq}{Nasdaq-100}
\newcommand{\totaleconreturn}{\nu}
\newcommand{\estavgtotaleconreturn}{\bar{\totaleconreturn}}
\newcommand{\tableenv}{env}

\newcommand{\matprSG}{mean annualized total portfolio return}
\newcommand{\matprPL}{mean annualized total portfolio returns}

\newcommand{\transfunc}{{\mathcal{P}}}

\newcommand{\stateset}{{\mathcal{S}}}
\newcommand{\observationset}{{\mathcal{O}}}

\newcommand{\actset}{{\mathcal{A}_{2C}}}

\newcommand{\actionset}{{\mathcal{A}}}
\newcommand{\actionsetconstrained}{\actionset}
\newcommand{\actionsetsurrogate}{\tilde{\actionset}}

\newcommand{\actionsurrogate}{\tilde{a}}
\newcommand{\actionconstrained}{a}

\newcommand{\policyconstrained}{\pi}
\newcommand{\policysurrogate}{\tilde{\pi}}

\newcommand{\economicreturnportfolio}{\Theta_{PF}}
\newcommand{\economicreturnportfoliorealized}{\vartheta_{PF}}

\newcommand{\economicreturnvector}{\Theta}
\newcommand{\economicreturnvectorrealized}{\vartheta}

\newcommand{\riskestfunc}{\hat{f}_{risk, \economicreturnportfolio}}
\newcommand{\riskfunc}{f_{risk, \economicreturnportfolio}}

\newcommand{\assetallocationset}{\mathcal{V}}
\newcommand{\wealthset}{\mathcal{W}}
\newcommand{\risktoleranceset}{\mathcal{T}}
\newcommand{\marketset}{\mathcal{U}}
\newcommand{\transactioncost}{tc}
\newcommand{\borrowingfee}{bf}
\newcommand{\E}{{\rm I\!E}}

\newcommand{\setupAshort}{(A) simulation}
\newcommand{\setupBshort}{(B) backtesting}
\newcommand{\setupA}{\setupAshort~setup}
\newcommand{\setupB}{\setupBshort~setup}

\newcommand{\numberexperiments}{100}

\appendix


\section{Asset Universe}
Table~\ref{tab:assetsenv} shows the assets from the \nasdaq~index used in the simulation environment and the backtesting environment.

\begin{table}
\centering
\begin{tabular}{C{0.5cm} | c C{1.2cm} C{2.5cm} }
 \toprule
 Index & ISIN & Ticker & Name \\
 \midrule
 0 & - & CASH & CASH\\
 1 & US7475251036 & QCOM & Qualcomm Inc.\\ 
 2 & US0567521085 & BIDU & Baidu Inc. \\
 3 & US0530151036 & ADP & Automatic Data Processing  Inc.\\
 4 & US8552441094 & SBUX & Starbucks Corporation\\
 5 & US0311621009 & AMGN & Amgen Inc.\\
 6 & US6937181088 & PCAR & Paccar Inc.\\
7 & US0231351067 & AMZN & Amazon.com Inc.\\
8 & US4612021034 & INTU & Intuit Inc.\\
9 & US00724F1012 & ADBE & Adobe Inc.\\
10 & US67066G1040 & NVDA & 
NVIDIA Corporation\\
11 & US5949181045 & MSFT & Microsoft Corporation\\
12 & US2786421030 & EBAY & eBay Inc.\\
 
\bottomrule
\end{tabular}
\caption{List of assets used in the environment.}
\label{tab:assetsenv}
\end{table}

\section{Constraint Configurations}
Tables~\ref{tab:constraintconfbeg}~-~\ref{tab:constraintconfend} show the constraint configurations used by the 100 different experiments.

\begin{table}
	\centering
	\begin{tabular}{C{1.8cm} C{3.6cm} C{1.4cm}}
		\toprule
		& $V_i$ (Affected Assets)& $c_i$ (Constraint Value)\\[0.5ex] 
		\midrule 
		\textbf{Experiment 0} & &\\
		Constraint 1& [CASH, BIDU, PCAR, NVDA, SBUX, INTU] & 0.4\\
		Constraint 2& [BIDU, NVDA, ADP] & 0.8\\
		\midrule
		\textbf{Experiment 1} & &\\
		Constraint 1& [BIDU, SBUX] & 0.42\\
		Constraint 2& [CASH, MSFT, BIDU, ADBE, NVDA, INTU, ADP, AMZN, EBAY, AMGN, QCOM] & 0.34\\
		\midrule
		\textbf{Experiment 2} & &\\
		Constraint 1& [BIDU, QCOM] & 0.23\\
		Constraint 2& [ADBE, SBUX, QCOM] & 0.32\\
		\midrule
		\textbf{Experiment 3} & &\\
		Constraint 1& [ADBE, SBUX, INTU, ADP, AMZN, AMGN, QCOM] & 0.86\\
		Constraint 2& [PCAR, ADP] & 0.28\\
		\midrule
		\textbf{Experiment 4} & &\\
		Constraint 1& [CASH, MSFT, ADBE, PCAR, QCOM] & 0.3\\
		Constraint 2& [BIDU, ADBE, NVDA, SBUX, EBAY] & 0.25\\
		\midrule
		\textbf{Experiment 5} & &\\
		Constraint 1& [CASH, PCAR, SBUX, AMGN] & 0.32\\
		Constraint 2& [CASH, MSFT, BIDU, ADBE, PCAR, NVDA, INTU, AMZN, EBAY, AMGN, QCOM] & 0.19\\
		\midrule
		\textbf{Experiment 6} & &\\
		Constraint 1& [CASH, BIDU, ADBE, NVDA, SBUX, INTU, ADP, EBAY, QCOM] & 0.87\\
		Constraint 2& [CASH, MSFT, ADBE, INTU, AMZN, EBAY] & 0.07\\
		\midrule
		\textbf{Experiment 7} & &\\
		Constraint 1& [CASH, BIDU, ADBE, PCAR, NVDA, SBUX, AMZN, EBAY, AMGN, QCOM] & 0.94\\
		Constraint 2& [CASH, MSFT, BIDU, PCAR, SBUX, ADP, AMZN, EBAY, QCOM] & 0.18\\
		\midrule
		\textbf{Experiment 8} & &\\
		Constraint 1& [MSFT, BIDU, PCAR, NVDA, INTU, QCOM] & 0.65\\
		Constraint 2& [CASH, BIDU, NVDA, INTU, AMGN] & 0.22\\
        \midrule
		\textbf{Experiment 9} & &\\
		Constraint 1& [CASH, MSFT, ADBE, PCAR, NVDA, SBUX, INTU, ADP, AMZN, AMGN, QCOM] & 0.39\\
		Constraint 2& [AMZN, AMGN] & 0.18\\
		\bottomrule
	\end{tabular}
	\caption{Details of Constraint Configurations for Experiments.}
	\label{tab:constraintconfbeg}
\end{table}

\begin{table}
	\centering
	\begin{tabular}{C{1.8cm} C{3.6cm} C{1.4cm}}
		\toprule
		& $V_i$ (Affected Assets)& $c_i$ (Constraint Value)\\[0.5ex] 
		\midrule 
		\textbf{Experiment 10} & &\\
		Constraint 1& [CASH, NVDA, INTU, EBAY] & 0.15\\
		Constraint 2& [CASH, BIDU, ADBE, PCAR, ADP, EBAY] & 0.36\\
		\midrule
		\textbf{Experiment 11} & &\\
		Constraint 1& [CASH, BIDU, ADBE, PCAR, SBUX, INTU, ADP, AMZN, EBAY, AMGN] & 0.89\\
		Constraint 2& [MSFT, BIDU, ADBE, SBUX, AMZN, AMGN] & 0.29\\
		\midrule
		\textbf{Experiment 12} & &\\
\bottomrule
	\end{tabular}
	\caption{Details of Constraint Configurations for Experiments.}
	\label{tab:evaluationinference}
\end{table}

\begin{table}
	\centering
	\begin{tabular}{C{1.8cm} C{3.6cm} C{1.4cm}}
		\toprule
		& $V_i$ (Affected Assets)& $c_i$ (Constraint Value)\\[0.5ex] 
		\midrule 
		\textbf{Experiment 20} & &\\
		Constraint 1& [CASH, ADBE, PCAR, NVDA, SBUX, AMZN] & 0.36\\
		Constraint 2& [NVDA, SBUX, INTU, ADP, AMGN, QCOM] & 0.73\\
		\midrule
		\textbf{Experiment 21} & &\\
		Constraint 1& [CASH, MSFT, BIDU, ADBE, PCAR, SBUX, ADP, AMZN, EBAY, QCOM] & 0.26\\
		Constraint 2& [CASH, MSFT, BIDU, ADBE, PCAR, NVDA, SBUX, INTU, AMZN, AMGN, QCOM] & 0.84\\
		\midrule
		\textbf{Experiment 22} & &\\
		Constraint 1& [BIDU, ADBE, SBUX, QCOM] & 0.33\\
		Constraint 2& [CASH, MSFT, BIDU, PCAR, INTU, AMZN] & 0.57\\
		\midrule
		\textbf{Experiment 23} & &\\
		Constraint 1& [MSFT, BIDU, PCAR, NVDA, ADP, AMZN, EBAY, QCOM] & 0.87\\
		Constraint 2& [MSFT, PCAR, NVDA, INTU, AMZN] & 0.48\\
		\midrule
		\textbf{Experiment 24} & &\\
		Constraint 1& [MSFT, BIDU, ADBE, AMZN, EBAY, AMGN] & 0.73\\
		Constraint 2& [CASH, MSFT, BIDU, ADBE, PCAR, SBUX, ADP, AMZN, EBAY, AMGN, QCOM] & 0.8\\
		\midrule
		\textbf{Experiment 25} & &\\
		Constraint 1& [CASH, BIDU, ADBE, PCAR, ADP, AMZN, EBAY] & 0.48\\
		Constraint 2& [CASH, MSFT, PCAR, SBUX, INTU, AMZN, AMGN, QCOM] & 0.68\\
		\midrule
		\textbf{Experiment 26} & &\\
		Constraint 1& [CASH, MSFT, ADBE, NVDA, SBUX, INTU, ADP, AMZN, QCOM] & 0.67\\
		Constraint 2& [CASH, MSFT, ADBE, NVDA, SBUX, ADP, AMZN] & 0.85\\
		\midrule
		\textbf{Experiment 27} & &\\
		Constraint 1& [CASH, ADBE, SBUX, ADP, AMGN] & 0.47\\
		Constraint 2& [CASH, MSFT, BIDU, ADBE, PCAR, SBUX, INTU, AMZN, EBAY, AMGN, QCOM] & 0.35\\
		\midrule
		\textbf{Experiment 28} & &\\
		Constraint 1& [CASH, PCAR, EBAY] & 0.23\\
		Constraint 2& [MSFT, BIDU, ADBE, PCAR, AMZN, EBAY, AMGN, QCOM] & 0.62\\
        \midrule
		\textbf{Experiment 29} & &\\
		Constraint 1& [SBUX, AMGN, QCOM] & 0.41\\
		Constraint 2& [MSFT, NVDA, SBUX, INTU, AMZN, AMGN, QCOM] & 0.37\\
		\bottomrule
	\end{tabular}
	\caption{Details of Constraint Configurations for Experiments.}
	\label{tab:evaluationinference}
\end{table}

\begin{table}
	\centering
	\begin{tabular}{C{1.8cm} C{3.6cm} C{1.4cm}}
		\toprule
		& $V_i$ (Affected Assets)& $c_i$ (Constraint Value)\\[0.5ex] 
		\midrule 
		\textbf{Experiment 30} & &\\
		Constraint 1& [CASH, MSFT, ADBE, PCAR, SBUX, INTU, ADP, AMZN, AMGN, QCOM] & 0.84\\
		Constraint 2& [ADBE, PCAR, INTU] & 0.24\\
		\midrule
		\textbf{Experiment 31} & &\\
		Constraint 1& [CASH, MSFT, PCAR, NVDA, AMZN] & 0.39\\
		Constraint 2& [NVDA, SBUX, INTU, AMZN, AMGN, QCOM] & 0.28\\
		\midrule
		\textbf{Experiment 32} & &\\
		Constraint 1& [ADBE, AMGN] & 0.45\\
		Constraint 2& [MSFT, BIDU, SBUX, INTU, ADP, EBAY, QCOM] & 0.4\\
		\midrule
		\textbf{Experiment 33} & &\\
		Constraint 1& [MSFT, ADBE, QCOM] & 0.74\\
		Constraint 2& [MSFT, PCAR, INTU, EBAY] & 0.09\\
		\midrule
		\textbf{Experiment 34} & &\\
		Constraint 1& [BIDU, PCAR, NVDA, SBUX, INTU, ADP, AMZN, EBAY, QCOM] & 0.95\\
		Constraint 2& [CASH, MSFT, BIDU, PCAR, NVDA, SBUX, EBAY, QCOM] & 0.53\\
		\midrule
		\textbf{Experiment 35} & &\\
		Constraint 1& [PCAR, ADP, AMGN] & 0.46\\
		Constraint 2& [NVDA, INTU, ADP, EBAY] & 0.07\\
		\midrule
		\textbf{Experiment 36} & &\\
		Constraint 1& [CASH, MSFT, BIDU, PCAR, NVDA, SBUX] & 0.93\\
		Constraint 2& [CASH, ADBE, PCAR, NVDA, AMZN, EBAY, AMGN] & 0.04\\
		\midrule
		\textbf{Experiment 37} & &\\
		Constraint 1& [CASH, BIDU, SBUX, INTU, AMGN, QCOM] & 0.04\\
		Constraint 2& [CASH, MSFT, PCAR, SBUX, ADP, EBAY] & 0.25\\
		\midrule
		\textbf{Experiment 38} & &\\
		Constraint 1& [NVDA, INTU, ADP, AMZN, AMGN] & 0.2\\
		Constraint 2& [BIDU, ADBE, INTU, AMZN] & 0.71\\
        \midrule
		\textbf{Experiment 39} & &\\
		Constraint 1& [MSFT, BIDU, ADBE, PCAR, NVDA, ADP, AMZN, EBAY, QCOM] & 0.22\\
		Constraint 2& [MSFT, ADBE, INTU, AMZN, EBAY, AMGN] & 0.42\\
		\bottomrule
	\end{tabular}
	\caption{Details of Constraint Configurations for Experiments.}
	\label{tab:evaluationinference}
\end{table}

\begin{table}
	\centering
	\begin{tabular}{C{1.8cm} C{3.6cm} C{1.4cm}}
		\toprule
		& $V_i$ (Affected Assets)& $c_i$ (Constraint Value)\\[0.5ex] 
		\midrule 
		\textbf{Experiment 40} & &\\
		Constraint 1& [PCAR, ADP, AMGN] & 0.18\\
		Constraint 2& [BIDU, ADBE, NVDA, SBUX, ADP, AMGN] & 0.55\\
		\midrule
		\textbf{Experiment 41} & &\\
		Constraint 1& [BIDU, SBUX, INTU, ADP, AMZN, EBAY, QCOM] & 0.92\\
		Constraint 2& [CASH, BIDU, PCAR, INTU, AMZN, AMGN, QCOM] & 0.78\\
		\midrule
		\textbf{Experiment 42} & &\\
\bottomrule
	\end{tabular}
	\caption{Details of Constraint Configurations for Experiments.}
	\label{tab:evaluationinference}
\end{table}

\begin{table}
	\centering
	\begin{tabular}{C{1.8cm} C{3.6cm} C{1.4cm}}
		\toprule
		& $V_i$ (Affected Assets)& $c_i$ (Constraint Value)\\[0.5ex] 
		\midrule 
		\textbf{Experiment 50} & &\\
		Constraint 1& [CASH, MSFT, BIDU, ADBE, NVDA, INTU, ADP, EBAY, AMGN, QCOM] & 0.05\\
		Constraint 2& [CASH, MSFT, ADBE, ADP, AMZN, QCOM] & 0.39\\
		\midrule
		\textbf{Experiment 51} & &\\
		Constraint 1& [CASH, MSFT, BIDU, ADBE, SBUX, INTU, EBAY, QCOM] & 0.3\\
		Constraint 2& [CASH, MSFT, BIDU, NVDA, SBUX, INTU, ADP, AMZN, EBAY, QCOM] & 0.24\\
		\midrule
		\textbf{Experiment 52} & &\\
		Constraint 1& [AMZN, AMGN, QCOM] & 0.03\\
		Constraint 2& [MSFT, NVDA, AMZN] & 0.94\\
		\midrule
		\textbf{Experiment 53} & &\\
		Constraint 1& [MSFT, BIDU, ADBE, PCAR, NVDA, SBUX, INTU, AMZN, EBAY, AMGN, QCOM] & 0.26\\
		Constraint 2& [CASH, MSFT, BIDU, ADBE, PCAR, SBUX, ADP, AMZN] & 0.02\\
		\midrule
		\textbf{Experiment 54} & &\\
		Constraint 1& [ADBE, PCAR, NVDA, SBUX, INTU, EBAY, AMGN] & 0.86\\
		Constraint 2& [MSFT, BIDU, ADBE, PCAR, SBUX, EBAY, AMGN, QCOM] & 0.86\\
		\midrule
		\textbf{Experiment 55} & &\\
		Constraint 1& [MSFT, ADP] & 0.87\\
		Constraint 2& [CASH, MSFT, QCOM] & 0.71\\
		\midrule
		\textbf{Experiment 56} & &\\
		Constraint 1& [MSFT, BIDU, PCAR, NVDA, INTU, ADP, AMGN, QCOM] & 0.0\\
		Constraint 2& [MSFT, BIDU, NVDA, SBUX, ADP, EBAY, AMGN, QCOM] & 0.56\\
		\midrule
		\textbf{Experiment 57} & &\\
		Constraint 1& [ADBE, NVDA, AMZN, EBAY] & 0.25\\
		Constraint 2& [CASH, PCAR, NVDA, SBUX, AMZN, AMGN] & 0.93\\
		\midrule
		\textbf{Experiment 58} & &\\
		Constraint 1& [CASH, MSFT, BIDU, ADBE, NVDA, SBUX, INTU, EBAY, QCOM] & 0.35\\
		Constraint 2& [CASH, MSFT, BIDU, ADBE, PCAR, NVDA, SBUX, INTU, ADP, AMZN, QCOM] & 0.68\\
        \midrule
		\textbf{Experiment 59} & &\\
		Constraint 1& [BIDU, ADBE, PCAR, SBUX, AMZN, QCOM] & 0.77\\
		Constraint 2& [BIDU, ADBE, NVDA, AMGN, QCOM] & 0.22\\
		\bottomrule
	\end{tabular}
	\caption{Details of Constraint Configurations for Experiments.}
	\label{tab:evaluationinference}
\end{table}

\begin{table}
	\centering
	\begin{tabular}{C{1.8cm} C{3.6cm} C{1.4cm}}
		\toprule
		& $V_i$ (Affected Assets)& $c_i$ (Constraint Value)\\[0.5ex] 
		\midrule 
		\textbf{Experiment 60} & &\\
		Constraint 1& [CASH, NVDA] & 0.87\\
		Constraint 2& [CASH, BIDU, ADBE, PCAR, NVDA, SBUX, INTU, ADP, EBAY, AMGN, QCOM] & 0.94\\
		\midrule
		\textbf{Experiment 61} & &\\
		Constraint 1& [SBUX, INTU, ADP, QCOM] & 0.0\\
		Constraint 2& [MSFT, INTU, EBAY, AMGN] & 0.39\\
		\midrule
		\textbf{Experiment 62} & &\\
		Constraint 1& [MSFT, BIDU, PCAR, SBUX, INTU, AMZN, EBAY, QCOM] & 0.68\\
		Constraint 2& [ADBE, PCAR, NVDA, ADP, AMZN, QCOM] & 0.24\\
		\midrule
		\textbf{Experiment 63} & &\\
		Constraint 1& [BIDU, INTU, ADP, QCOM] & 0.57\\
		Constraint 2& [ADBE, AMZN] & 0.34\\
		\midrule
		\textbf{Experiment 64} & &\\
		Constraint 1& [CASH, MSFT, PCAR, NVDA, SBUX, INTU, AMZN, EBAY, AMGN, QCOM] & 0.41\\
		Constraint 2& [CASH, MSFT, ADBE, PCAR, NVDA, SBUX, INTU, ADP, AMZN, EBAY, QCOM] & 0.83\\
		\midrule
		\textbf{Experiment 65} & &\\
		Constraint 1& [INTU, ADP] & 0.27\\
		Constraint 2& [ADBE, PCAR, SBUX, AMZN] & 0.55\\
		\midrule
		\textbf{Experiment 66} & &\\
		Constraint 1& [BIDU, ADBE, PCAR, SBUX, INTU, AMZN, EBAY, AMGN, QCOM] & 0.22\\
		Constraint 2& [MSFT, BIDU, PCAR, NVDA, SBUX, INTU, EBAY, AMGN, QCOM] & 0.98\\
		\midrule
		\textbf{Experiment 67} & &\\
		Constraint 1& [MSFT, NVDA] & 0.19\\
		Constraint 2& [CASH, BIDU, NVDA, INTU, AMZN] & 0.49\\
		\midrule
		\textbf{Experiment 68} & &\\
		Constraint 1& [MSFT, PCAR, SBUX, INTU, AMZN, EBAY] & 0.65\\
		Constraint 2& [MSFT, BIDU, ADBE, NVDA, SBUX, INTU, AMZN, EBAY] & 0.07\\
        \midrule
		\textbf{Experiment 69} & &\\
		Constraint 1& [CASH, ADBE, PCAR, SBUX, ADP, AMZN, EBAY, AMGN, QCOM] & 0.94\\
		Constraint 2& [NVDA, AMZN] & 0.45\\
		\bottomrule
	\end{tabular}
	\caption{Details of Constraint Configurations for Experiments.}
	\label{tab:evaluationinference}
\end{table}

\begin{table}
	\centering
	\begin{tabular}{C{1.8cm} C{3.6cm} C{1.4cm}}
		\toprule
		& $V_i$ (Affected Assets)& $c_i$ (Constraint Value)\\[0.5ex] 
		\midrule 
		\textbf{Experiment 70} & &\\
		Constraint 1& [CASH, MSFT, BIDU, ADBE, SBUX, INTU, ADP, EBAY, AMGN, QCOM] & 0.57\\
		Constraint 2& [MSFT, PCAR, NVDA, SBUX, ADP, AMZN, EBAY, AMGN, QCOM] & 0.26\\
		\midrule
		\textbf{Experiment 71} & &\\
		Constraint 1& [CASH, MSFT, BIDU, NVDA, INTU, AMZN, EBAY] & 0.56\\
		Constraint 2& [CASH, PCAR, NVDA, SBUX, INTU, ADP, QCOM] & 0.3\\
		\midrule
		\textbf{Experiment 72} & &\\
\bottomrule
	\end{tabular}
	\caption{Details of Constraint Configurations for Experiments.}
	\label{tab:evaluationinference}
\end{table}

\begin{table}
	\centering
	\begin{tabular}{C{1.8cm} C{3.6cm} C{1.4cm}}
		\toprule
		& $V_i$ (Affected Assets)& $c_i$ (Constraint Value)\\[0.5ex] 
		\midrule 
		\textbf{Experiment 80} & &\\
		Constraint 1& [CASH, MSFT, BIDU, ADBE, PCAR, SBUX, QCOM] & 0.74\\
		Constraint 2& [MSFT, BIDU, NVDA, AMZN, EBAY] & 0.86\\
		\midrule
		\textbf{Experiment 81} & &\\
		Constraint 1& [CASH, MSFT, BIDU, ADBE, NVDA, SBUX, INTU, ADP, AMZN, QCOM] & 0.68\\
		Constraint 2& [MSFT, BIDU, PCAR, AMZN, AMGN] & 0.13\\
		\midrule
		\textbf{Experiment 82} & &\\
		Constraint 1& [CASH, MSFT, BIDU] & 0.49\\
		Constraint 2& [CASH, BIDU, PCAR, SBUX, INTU, EBAY, QCOM] & 0.66\\
		\midrule
		\textbf{Experiment 83} & &\\
		Constraint 1& [PCAR, NVDA, INTU, AMZN] & 0.15\\
		Constraint 2& [CASH, MSFT, ADBE, PCAR, NVDA, SBUX, AMZN, AMGN] & 0.91\\
		\midrule
		\textbf{Experiment 84} & &\\
		Constraint 1& [MSFT, BIDU, ADBE, NVDA, SBUX, ADP, AMZN, AMGN, QCOM] & 0.0\\
		Constraint 2& [ADBE, PCAR, AMZN, EBAY] & 0.87\\
		\midrule
		\textbf{Experiment 85} & &\\
		Constraint 1& [ADBE, SBUX, ADP] & 0.32\\
		Constraint 2& [CASH, MSFT, BIDU, INTU, ADP, AMZN, EBAY, AMGN] & 0.36\\
		\midrule
		\textbf{Experiment 86} & &\\
		Constraint 1& [CASH, MSFT, EBAY, AMGN] & 0.55\\
		Constraint 2& [CASH, ADBE, NVDA, SBUX, INTU, ADP, AMZN, EBAY, AMGN, QCOM] & 0.07\\
		\midrule
		\textbf{Experiment 87} & &\\
		Constraint 1& [MSFT, BIDU, PCAR, NVDA, INTU, ADP, AMZN, AMGN, QCOM] & 0.82\\
		Constraint 2& [CASH, MSFT, ADBE, PCAR, ADP, AMZN, EBAY, QCOM] & 0.48\\
		\midrule
		\textbf{Experiment 88} & &\\
		Constraint 1& [CASH, BIDU, ADBE, PCAR, NVDA, ADP, AMZN, EBAY, AMGN] & 0.28\\
		Constraint 2& [CASH, AMZN] & 0.66\\
        \midrule
		\textbf{Experiment 89} & &\\
		Constraint 1& [CASH, ADBE, PCAR, NVDA, SBUX, INTU, ADP, AMZN, EBAY, AMGN, QCOM] & 0.02\\
		Constraint 2& [CASH, MSFT, BIDU, ADBE, NVDA, INTU, ADP, AMGN, QCOM] & 0.2\\
		\bottomrule
	\end{tabular}
	\caption{Details of Constraint Configurations for Experiments.}
	\label{tab:evaluationinference}
\end{table}

\begin{table}
	\centering
	\begin{tabular}{C{1.8cm} C{3.6cm} C{1.4cm}}
		\toprule
		& $V_i$ (Affected Assets)& $c_i$ (Constraint Value)\\[0.5ex] 
		\midrule 
		\textbf{Experiment 90} & &\\
		Constraint 1& [CASH, MSFT, BIDU, ADBE, PCAR, NVDA, SBUX, ADP, AMZN, AMGN, QCOM] & 0.07\\
		Constraint 2& [CASH, MSFT, BIDU, ADBE, PCAR, NVDA, INTU, AMZN, EBAY, QCOM] & 0.43\\
		\midrule
		\textbf{Experiment 91} & &\\
		Constraint 1& [INTU, ADP] & 0.95\\
		Constraint 2& [ADBE, PCAR, NVDA, SBUX, ADP, AMZN, AMGN, QCOM] & 0.15\\
		\midrule
		\textbf{Experiment 92} & &\\
		Constraint 1& [BIDU, ADBE, NVDA, SBUX, INTU, ADP, AMZN, EBAY, AMGN, QCOM] & 0.04\\
		Constraint 2& [MSFT, BIDU, PCAR, NVDA, ADP, AMZN, AMGN, QCOM] & 0.79\\
		\midrule
		\textbf{Experiment 93} & &\\
		Constraint 1& [ADBE, SBUX, ADP] & 0.28\\
		Constraint 2& [MSFT, BIDU, PCAR, NVDA, SBUX, AMZN, AMGN, QCOM] & 0.73\\
		\midrule
		\textbf{Experiment 94} & &\\
		Constraint 1& [MSFT, ADBE, PCAR, SBUX, ADP, AMZN, EBAY, QCOM] & 0.3\\
		Constraint 2& [ADBE, SBUX, INTU, EBAY] & 0.14\\
		\midrule
		\textbf{Experiment 95} & &\\
		Constraint 1& [BIDU, NVDA, AMZN, AMGN, QCOM] & 0.92\\
		Constraint 2& [CASH, MSFT, BIDU, PCAR, NVDA, SBUX, INTU, ADP, AMZN, EBAY, AMGN] & 0.91\\
		\midrule
		\textbf{Experiment 96} & &\\
		Constraint 1& [BIDU, ADBE, PCAR, NVDA, SBUX, INTU, EBAY, AMGN] & 0.05\\
		Constraint 2& [MSFT, BIDU, ADBE, NVDA, SBUX, INTU, ADP, AMZN, EBAY, AMGN, QCOM] & 0.34\\
		\midrule
		\textbf{Experiment 97} & &\\
		Constraint 1& [EBAY, QCOM] & 0.61\\
		Constraint 2& [CASH, MSFT, BIDU, ADBE, PCAR, NVDA, SBUX, INTU, AMZN, AMGN, QCOM] & 0.85\\
		\midrule
		\textbf{Experiment 98} & &\\
		Constraint 1& [MSFT, INTU, EBAY, AMGN] & 0.65\\
		Constraint 2& [MSFT, NVDA, INTU, AMGN] & 0.67\\
        \midrule
		\textbf{Experiment 99} & &\\
		Constraint 1& [CASH, MSFT, ADBE] & 0.88\\
		Constraint 2& [CASH, ADBE, PCAR, INTU, AMZN, EBAY] & 0.18\\
		\bottomrule
	\end{tabular}
	\caption{Details of Constraint Configurations for Experiments.}
	\label{tab:constraintconfend}
\end{table}

\section{Hyperparameters}

Tables~\ref{tab:hyperparambeg}~-~\ref{tab:hyperparamend} show the hyperparameters used for each approach in the 100 experiments after hyperparameter tuning was applied.

\begin{table}
	\centering
	\begin{tabular}{C{1.2cm} C{0.5cm} C{0.7cm} C{0.8cm} C{0.6cm} C{0.5cm} C{1.8cm}}
		\toprule
	    \textbf{RCPO}	& seed & learning rate & lambda learning rate & batch size & mini batch size & hidden layer size\\[0.5ex]
	    \midrule
        Expt. 0 & 149 & 5e-06 & 0.0005 & 4000 & 128 & [512, 256, 128]\\
		Expt. 1 & 470 & 5e-06 & 5e-05 & 4000 & 128 & [512, 256, 128]\\
		Expt. 2 & 465 & 5e-06 & 0.0005 & 4000 & 128 & [512, 256, 128]\\
		Expt. 3 & 974 & 5e-06 & 0.0005 & 4000 & 128 & [512, 256, 128]\\
		Expt. 4 & 789 & 5e-06 & 5e-05 & 4000 & 128 & [512, 256, 128]\\
		Expt. 5 & 178 & 5e-06 & 5e-05 & 4000 & 128 & [512, 256, 128]\\
		Expt. 6 & 722 & 5e-06 & 5e-05 & 4000 & 128 & [512, 256, 128]\\
\bottomrule
	\end{tabular}
	\caption{Hyperparameter settings for RCPO.}
	\label{tab:hyperparambeg}
\end{table}
\begin{table}
	\centering
	\begin{tabular}{C{1.2cm} C{0.5cm} C{0.7cm} C{0.8cm} C{0.6cm} C{0.5cm} C{1.8cm}}
		\toprule
	    \textbf{RCPO}	& seed & learning rate & lambda learning rate & batch size & mini batch size & hidden layer size\\[0.5ex]
	    \midrule
        Expt. 50 & 202 & 5e-06 & 0.0005 & 4000 & 128 & [512, 256, 128]\\
		Expt. 51 & 956 & 5e-06 & 0.0001 & 4000 & 128 & [512, 256, 128]\\
		Expt. 52 & 254 & 5e-06 & 5e-05 & 4000 & 128 & [512, 256, 128]\\
		Expt. 53 & 161 & 5e-06 & 0.001 & 4000 & 128 & [512, 256, 128]\\
		Expt. 54 & 96 & 5e-06 & 5e-05 & 4000 & 128 & [512, 256, 128]\\
		Expt. 55 & 39 & 5e-06 & 5e-05 & 4000 & 128 & [512, 256, 128]\\
		Expt. 56 & 143 & 5e-06 & 0.001 & 4000 & 128 & [512, 256, 128]\\
\bottomrule
	\end{tabular}
	\caption{Hyperparameter settings for RCPO.}
	\label{tab:appendix}
\end{table}

\begin{table}
	\centering
	\begin{tabular}{C{1.2cm} C{0.5cm} C{0.7cm} C{0.8cm} C{0.6cm} C{0.5cm} C{1.8cm}}
		\toprule
	    \textbf{IPO}	& seed & learning rate & t weight & batch size & mini batch size & hidden layer size\\[0.5ex]
	    \midrule
        Expt. 0 & 811 & 5e-06 & 20 & 4000 & 128 & [512, 256, 128]\\
		Expt. 1 & 837 & 5e-06 & 10 & 4000 & 128 & [512, 256, 128]\\
		Expt. 2 & 384 & 1e-05 & 20 & 4000 & 128 & [512, 256, 128]\\
		Expt. 3 & 164 & 5e-06 & 10 & 4000 & 128 & [512, 256, 128]\\
		Expt. 4 & 877 & 5e-06 & 10 & 4000 & 128 & [512, 256, 128]\\
		Expt. 5 & 664 & 1e-05 & 10 & 4000 & 128 & [512, 256, 128]\\
		Expt. 6 & 653 & 5e-06 & 10 & 4000 & 128 & [512, 256, 128]\\
\bottomrule
	\end{tabular}
	\caption{Hyperparameter settings for IPO.}
	\label{tab:appendix}
\end{table}
\begin{table}
	\centering
	\begin{tabular}{C{1.2cm} C{0.5cm} C{0.7cm} C{0.8cm} C{0.6cm} C{0.5cm} C{1.8cm}}
		\toprule
	    \textbf{IPO}	& seed & learning rate & t weight & batch size & mini batch size & hidden layer size\\[0.5ex]
	    \midrule
        Expt. 50 & 399 & 1e-05 & 10 & 4000 & 128 & [512, 256, 128]\\
		Expt. 51 & 977 & 5e-06 & 10 & 4000 & 128 & [512, 256, 128]\\
		Expt. 52 & 277 & 5e-06 & 10 & 4000 & 128 & [512, 256, 128]\\
		Expt. 53 & 2 & 1e-05 & 10 & 4000 & 128 & [512, 256, 128]\\
		Expt. 54 & 277 & 5e-06 & 10 & 4000 & 128 & [512, 256, 128]\\
		Expt. 55 & 815 & 1e-05 & 10 & 4000 & 128 & [512, 256, 128]\\
		Expt. 56 & 311 & 1e-05 & 10 & 4000 & 128 & [512, 256, 128]\\
\bottomrule
	\end{tabular}
	\caption{Hyperparameter settings for IPO.}
	\label{tab:appendix}
\end{table}

\begin{table}
	\centering
	\begin{tabular}{C{1.2cm} C{0.5cm} C{0.7cm} C{0.8cm} C{0.6cm} C{0.5cm} C{1.8cm}}
		\toprule
	    \textbf{P3O} & seed & learning rate & init $\kappa$ & batch size & mini batch size & hidden layer size\\[0.5ex]
	    \midrule
        Expt. 0 & 698 & 1e-05 & 50 & 4000 & 128 & [512, 256, 128]\\
		Expt. 1 & 351 & 5e-06 & 20 & 4000 & 128 & [512, 256, 128]\\
		Expt. 2 & 555 & 1e-05 & 50 & 4000 & 128 & [512, 256, 128]\\
		Expt. 3 & 874 & 1e-05 & 50 & 4000 & 128 & [512, 256, 128]\\
		Expt. 4 & 965 & 5e-06 & 50 & 4000 & 128 & [512, 256, 128]\\
		Expt. 5 & 371 & 5e-06 & 20 & 4000 & 128 & [512, 256, 128]\\
		Expt. 6 & 495 & 1e-05 & 50 & 4000 & 128 & [512, 256, 128]\\
\bottomrule
	\end{tabular}
	\caption{Hyperparameter settings for P30.}
	\label{tab:appendix}
\end{table}
\begin{table}
	\centering
	\begin{tabular}{C{1.2cm} C{0.5cm} C{0.7cm} C{0.8cm} C{0.6cm} C{0.5cm} C{1.8cm}}
		\toprule
	    \textbf{P3O} & seed & learning rate & init $\kappa$ & batch size & mini batch size & hidden layer size\\[0.5ex]
	    \midrule
        Expt. 50 & 162 & 5e-06 & 50 & 4000 & 128 & [512, 256, 128]\\
		Expt. 51 & 912 & 1e-05 & 20 & 4000 & 128 & [512, 256, 128]\\
		Expt. 52 & 713 & 5e-06 & 20 & 4000 & 128 & [512, 256, 128]\\
		Expt. 53 & 3 & 5e-06 & 50 & 4000 & 128 & [512, 256, 128]\\
		Expt. 54 & 698 & 5e-06 & 50 & 4000 & 128 & [512, 256, 128]\\
		Expt. 55 & 752 & 5e-06 & 20 & 4000 & 128 & [512, 256, 128]\\
		Expt. 56 & 13 & 5e-06 & 20 & 4000 & 128 & [512, 256, 128]\\
\bottomrule
	\end{tabular}
	\caption{Hyperparameter settings for P30.}
	\label{tab:appendix}
\end{table}

\begin{table}
	\centering
	\begin{tabular}{C{1.2cm} C{0.5cm} C{0.8cm} C{0.6cm} C{0.5cm} C{1.8cm}}
		\toprule
	    \textbf{CAOSD} & seed & learning rate & batch size & mini batch size & hidden layer size\\[0.5ex]
	    \midrule
        Expt. 0 & 224 & 1e-06 & 4000 & 128 & [512, 256, 128]\\
		Expt. 1 & 907 & 1e-06 & 4000 & 128 & [512, 256, 128]\\
		Expt. 2 & 317 & 1e-06 & 4000 & 128 & [512, 256, 128]\\
		Expt. 3 & 537 & 1e-06 & 4000 & 128 & [512, 256, 128]\\
		Expt. 4 & 636 & 1e-06 & 4000 & 128 & [512, 256, 128]\\
		Expt. 5 & 221 & 1e-06 & 4000 & 128 & [512, 256, 128]\\
		Expt. 6 & 666 & 1e-06 & 4000 & 128 & [512, 256, 128]\\
\bottomrule
	\end{tabular}
	\caption{Hyperparameter settings for CAOSD.}
	\label{tab:appendix}
\end{table}
\begin{table}
	\centering
	\begin{tabular}{C{1.2cm} C{0.5cm} C{0.8cm} C{0.6cm} C{0.5cm} C{1.8cm}}
		\toprule
	    \textbf{CAOSD} & seed & learning rate & batch size & mini batch size & hidden layer size\\[0.5ex]
	    \midrule
        Expt. 50 & 440 & 1e-06 & 4000 & 128 & [512, 256, 128]\\
		Expt. 51 & 725 & 1e-06 & 4000 & 128 & [512, 256, 128]\\
		Expt. 52 & 359 & 1e-06 & 4000 & 128 & [512, 256, 128]\\
		Expt. 53 & 881 & 1e-06 & 4000 & 128 & [512, 256, 128]\\
		Expt. 54 & 990 & 1e-06 & 4000 & 128 & [512, 256, 128]\\
		Expt. 55 & 89 & 1e-06 & 4000 & 128 & [512, 256, 128]\\
		Expt. 56 & 71 & 0.0005 & 4000 & 128 & [512, 256, 128]\\
\bottomrule
	\end{tabular}
	\caption{Hyperparameter settings for CAOSD.}
	\label{tab:hyperparamend}
\end{table}

\bibliography{ecai}